\pdfoutput=1

\documentclass[11pt]{article}
\usepackage[table]{xcolor}
\usepackage{acl}

\usepackage{times}
\usepackage{latexsym}
\usepackage{algorithm}
\usepackage{algpseudocodex}
\usepackage{amssymb,amsmath,amsthm,enumitem}
\usepackage[capitalize]{cleveref} 
\usepackage{bbm}
\usepackage{inconsolata}
\usepackage{float}
\usepackage{fancyhdr}
\usepackage{mathtools}
\usepackage[T1]{fontenc}

\usepackage[utf8]{inputenc}
\usepackage{booktabs,tabularx,enumitem,ragged2e}
\usepackage[skins,breakable]{tcolorbox}
\usepackage{microtype}

\usepackage{inconsolata}
\usepackage{xcolor}
\usepackage{footmisc}
\crefname{section}{Sec.}{Secs.}
\usepackage{color, colortbl}
\usepackage{listings,multicol}
\colorlet{shadecolor}{gray!10}
\usepackage{tikz}

\newcommand{\execfig}{\includegraphics[scale=0.12,trim={0 2em 0 0}]{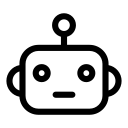}}%
\newcommand{\planfig}{\includegraphics[scale=0.175,trim={2em 3em 2em 0}]{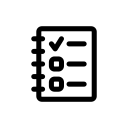}}%
\newcommand{\controlfig}{\includegraphics[scale=0.13,trim={0 2em 0em 0}]{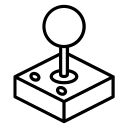}}%

\newcommand{\edited}[1]{\textcolor{black}{#1}}

\algrenewcommand\algorithmicrequire{\textbf{Input:}}
\algrenewcommand\algorithmicensure{\textbf{Output:}}
\algnewcommand{\IfThen}[2]{
  \State \algorithmicif\ #1\ \algorithmicthen\ #2}
\newcommand{\method}[0]{\textsc{ADaPT}}
\newcommand{\fullform}[0]{\textbf{A}s-Needed \textbf{D}ecomposition \textbf{a}nd \textbf{P}lanning for complex \textbf{T}asks}

\definecolor{comm}{gray}{0.6}
\definecolor{lightgray}{rgb}{0.83, 0.83, 0.83}
\definecolor{light-gray}{gray}{0.9}
\definecolor{darkturquoise}{rgb}{0.0, 0.81, 0.82}
\definecolor{dodgerblue}{rgb}{0.12, 0.56, 1.0}
\definecolor{lavender}{rgb}{0.9, 0.9, 0.98}
\definecolor{lightmauve}{rgb}{0.86, 0.82, 1.0}
\definecolor{peach}{rgb}{1.0, 0.94, 0.84}
\usepackage{graphicx}
\usepackage{multirow}
\title{
\textsc{ADaPT}: As-Needed Decomposition and Planning with Language Models}

\author{Archiki Prasad$^{\clubsuit}$ 
\quad Alexander Koller$^{\heartsuit}$ \quad Mareike Hartmann$^{\heartsuit}$ \vspace{2pt} \\
\quad \textbf{Peter Clark}$^{\spadesuit}$   \quad \textbf{Ashish Sabharwal}$^{\spadesuit}$ \quad \textbf{Mohit Bansal}$^{\clubsuit}$ \quad \textbf{Tushar Khot}$^{\spadesuit}$  \vspace{8pt} \\
$^{\clubsuit}$ UNC Chapel Hill \; $^{\spadesuit}$ Allen Institute for AI \; $^{\heartsuit}$ Saarland University
}

\begin{document}
\maketitle
\begin{abstract}

Large Language Models (LLMs) are increasingly being used for interactive decision-making tasks requiring planning and adapting to the environment.
Recent works employ LLMs-as-agents in broadly two ways:  iteratively determining the next action (iterative executors) or generating plans and executing sub-tasks using LLMs (plan-and-execute).
However, these methods struggle with task complexity, as the inability to execute any sub-task may lead to  task failure. 
To address these shortcomings, we introduce \fullform{} (\method{}), an approach that explicitly plans and decomposes complex sub-tasks \textit{as-needed}, i.e., when the LLM is unable to execute them.
\method{} recursively decomposes sub-tasks to adapt to both task complexity and LLM capability. 
 Our results demonstrate that \method{} substantially outperforms established strong baselines, achieving success rates up to $28.3\%$ higher in ALFWorld, $27\%$ in WebShop, and $33\%$ in TextCraft -- a novel compositional dataset that we introduce. Through extensive analysis, we illustrate the importance of multi-level decomposition and establish that \method{} dynamically adjusts to the capabilities of the executor LLM as well as to task complexity.\footnote{
 Project: \url{https://allenai.github.io/adaptllm}
 }
 
\end{abstract}
\begin{figure}[t]
    \centering
    \includegraphics[width=\linewidth]{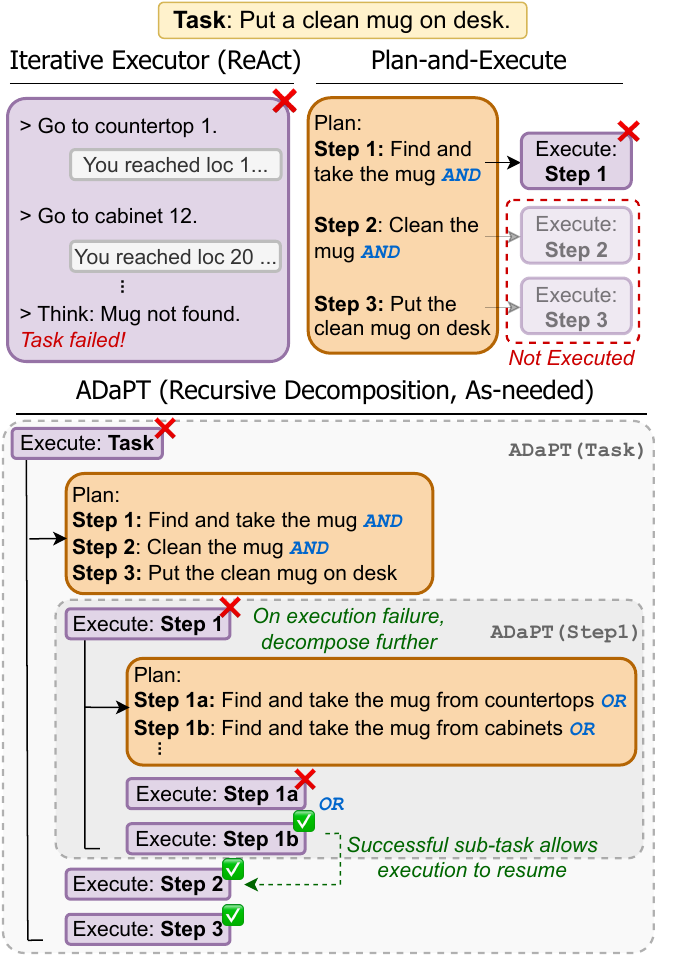}
    \caption{
    \textbf{Top-Left: }Iterative executors such as ReAct~\cite{yao2023react} interact directly with the environment, performing planning implicitly. \textbf{Top-Right:} Plan-and-Execute, e.g., \citet{yang2023intercode}, creates a fixed plan for the task, without accounting for complexity in executing step 1. \textbf{Bottom:} \method{} dynamically decomposes based on success of the executor. 
    }
    \label{fig:intro}
\end{figure}
\section{Introduction}
\label{sec:intro}

Recent advances in Large Language Models (LLMs) have expanded their application beyond conventional NLP tasks to more complex tasks involving mathematical, symbolic, and commonsense reasoning~\cite{wei2022chain,huang-chang-2023-towards}. Recent models have even been applied to \textit{decision-making} tasks, such as performing household chores, navigating a webpage, etc., that require interactions with external environments or tools~\cite{yao2023react, qin2023toolllm}.

Prior works on using LLMs for decision-making, such as ReAct~\cite{yao2023react}, iteratively generate the next action to be executed in the environment given the history of actions and observations (see~\cref{fig:intro}; top-left). However, as the tasks become more complex, LLMs struggle due to their limited composition ability~\cite{dziri2023faith} and inability to deal with the distractors~\cite{Shi2023LargeLM} in a long action-observation trajectory. 

To mitigate this, modular approaches~\cite{khot2023decomposed, yang2023intercode, Sun2023PEARLPL}
incorporate a separate planner module that utilizes an 
LLM to create a high-level plan.\footnote{By ``planning'', we refer to the colloquial concept of designing a list of sub-tasks to accomplish a complex task rather than its usage in classical AI-planning literature. E.g., a ``plan'' for preparing a lasagna could be to cook the pasta, prepare the sauce, layer the ingredients, and then bake it. 
}  The planner then delegates simpler sub-tasks  to an executor LLM module thereby reducing the compositional complexity and length of action trajectory required by the executor. We refer to this category broadly as \emph{plan-and-execute} approaches (see \cref{fig:intro}; top-right). While the plans enable these methods to guide the execution and track progress~\cite{wang-etal-2023-plan}, their non-adaptive nature poses a limitation when confronting unachievable sub-tasks. These approaches inherently lack the flexibility to adapt to task complexity and manage execution failures, as shown in \cref{fig:intro}(top-right), where just one sub-task that is too complex results in overall task failure.

To address such failures, we propose \fullform{} (\method{}),
a recursive algorithm that further decomposes sub-tasks \emph{when necessary}, to dynamically accommodate to task complexity. 
We utilize separate \emph{planner} and \emph{executor} LLM modules within our framework but \emph{only} decompose a task using the planner, if the executor LLM detects a failure. 
As shown in \cref{fig:intro}, the overall task of putting a clean mug on a desk in an unfamiliar household is too complex for the model, leading to failure of the iterative executor. While a plan-and-execute-style approach 
 initially breaks down the task into three sub-tasks, it falls short 
in accounting for the complexity in finding a mug.
 Moreover, it is challenging to anticipate the difficulty of such a sub-task in advance, as the executor could find a mug in the first attempt or in an obscure location.
Therefore, \method{} employs its recursive structure to \emph{dynamically adapt} to execution failures (assessed by LLMs), by \emph{further decomposing} the complex sub-task of \emph{finding a mug} via the planner.

Empirically, we demonstrate the effectiveness of \method{} on three datasets involving interactive environments: ALFWorld~\cite{sridhar2021alfworld}, WebShop~\cite{yao2022webshop}, and a new compositional text game for crafting Minecraft recipes called \emph{TextCraft} (\cref{ssec:data}). 
 Using GPT-3.5 as the underlying LLM, \method{} outperforms strong baselines (discussed in \cref{ssec:baselines}) such as ReAct~\cite{yao2023react}, and Plan-and-Solve~\cite{wang-etal-2023-plan}
 by up to $28.3\%$, $27\%$, and $33\%$ absolute points on ALFWorld, WebShop, and TextCraft respectively~(\cref{sec:result}). Compared to Reflexion~\cite{shinn2023reflexion}, an adaptive approach that addresses \emph{failures in the full task trajectory}, \method{} yields higher success rates by $14.1\%$, $9\%$, and $20\%$ on ALFWorld, WebShop, and TextCraft respectively.
 Through extensive analysis of \method{}, we establish the importance of recursive decomposition~(\cref{ssec:rq1}) and showcase dynamic adaptation to the capabilities of the executor LLM including open-source models such LLaMA-2~\cite{touvron2023llama} and Lemur~\cite{xu2023lemur} in \cref{ssec:rq2}.
 Lastly, we demonstrate that \method{} incorporates task complexity~(\cref{ssec:rq3}), where the extent of recursive decomposition aligns with the inherent task complexity.
To summarize, our contributions are:

\begin{enumerate}[leftmargin=*,noitemsep,nolistsep]

\item  We present \method{}, a recursive algorithm that dynamically decomposes complex sub-tasks on an as-needed basis, i.e., \emph{intervening only if the task is too complex for the executor}.

\item On three diverse datasets, ALFWorld, WebShop, and TextCraft, \method{} improves success rate of GPT-3.5 over previous approaches by up to $28.3\%$, $27\%$, and $33\%$ points respectively.

\item Analysis of \method{} underscores the significance of recursive decomposition and the ability to adapt dynamically to varying LLM execution capabilities and task complexities.

\end{enumerate}

\section{Related Work}
\label{sec:rel}

\paragraph{LLMs for Decision-Making.} LLMs have been successfully used as agents to perform a wide variety of decision-making tasks such as robotic navigation~\cite{ahn2022can, huang2023inner, singh2023progprompt}, complex multi-modal games like Minecraft~\cite{fan2022minedojo, wang2023voyager}, text-based environments~\cite{sridhar2021alfworld, liu2023agentbench}. While most of these works focus on learning from trajectories,  ReAct~\cite{yao2023react} uses few-shot prompting to build an agent that reasons about the current state (thoughts) and generates the next action in the environment, given prior actions and observations. Their iterative approach (shown in \cref{fig:intro}; top-left) can handle failures, but
 they have to keep track of the entire plan \emph{implicitly} while deciding every local action (c.f. \method{} in \cref{fig:alf_roll} of \cref{app:details}). By incorporating planning and execution into separate modules and enabling dynamic adaptation we are able to achieve higher success rates (refer to \cref{sec:result}).

Several follow-up works improve upon the ReAct framework by incorporating feedback in future trials~\cite{madaan2023self, shinn2023reflexion}, or using LLMs to develop heuristics for search~\cite{yao2023tree, zhou2023language}. In contrast to \method{}, they do not employ task decomposition, leading to unnecessary computation as they explore multiple trajectories or trials for the whole task, even though the LLM struggles with just one sub-task. Such works are complementary to \method{} as they can be incorporated within the planner or executor modules to strengthen LLM performance (just like they are incorporated in ReAct).

\paragraph{Decomposition and Modularity.} Our work follows extensive literature in NLP on decomposing tasks into neural modules ~\cite{andreas2016neural, gupta2019neural, jiang-bansal-2019-self} or seq2seq models~\cite{min-etal-2019-multi,talmor-berant-2018-web,khot-etal-2021-text,perez-etal-2020-unsupervised, saha2022summarization}. With the advent of few-shot prompted black-box LLMs, this paradigm of programmatic decomposition into LLMs has become more popular~\cite[\textit{inter alia}]{yao2023react,khot2023decomposed,wang-etal-2023-plan}, referred to as LLM Programs~\cite{schlag2023large,dohan2022language}. 
\edited{
Additionally, past works in program synthesis~\cite{murali2018neural,nye2019learning,zheng2023outline} also employ task decomposition via generating a ``program sketch'' prior to program generation.
}

\method{} not only decomposes tasks via the planner module and delegates them to the executor module but also \emph{automatically} adapts to executor failures by further decomposing complex tasks \emph{as-needed}. This dynamic capability distinguishes \method{} from prior works 
with a non-adaptive structure. 
\method{} extends the recursive and hierarchical decomposition in \citet{khot2023decomposed}, enabling inter-module communications, and robust strategies for execution failures, excelling in real-world textual environments like online shopping. 
 
\paragraph{Hierarchical Problem Solving.} 
In AI problem-solving, there is a longstanding tradition of hierarchical task decomposition employed in planning \cite{ghallab2004automated,georgievski2014overview,holler2020hddl}, reinforcement learning \cite{sutton1999between,  barto2003recent, nachum2018data, zhang2021hierarchical}, and navigation~\cite{she2014back, sharma-etal-2022-skill,  blukis2022persistent, min2022film, song2023llm}. These approaches, such as Hierarchical Task Networks~\cite{erol1994htn}, leverage domain knowledge, e.g., hand-specified library of plans, to break complex problems into simpler tasks. 
Our work embraces this tradition but distinguishes itself by exploring how LLMs can autonomously decompose tasks by leveraging their extensive world knowledge, without predefined plan libraries.  Lastly, \method{} performs dynamic hierarchical planning by employing its recursive structure. 

\section{Methodology}
\label{sec:method}
We introduce \fullform{} (\method{}), a modular approach for decision-making  that integrates an LLM as an \emph{executor} and a \emph{planner} (\cref{ssec:exec,ssec:plan}) within an LLM program called the controller (\cref{ssec:cont}). In \cref{fig:intro}, when \method{} is given a complex task, it first attempts to accomplish the entire task by running the executor iteratively, 
and resorting to the LLM planner for further decomposition into sub-tasks if the executor fails.
Subsequently, \method{} is recursively called for each sub-task to ensure their successful completion, ultimately leading to overall task success.

\begin{figure*}
    \centering
    \includegraphics[scale=0.57]{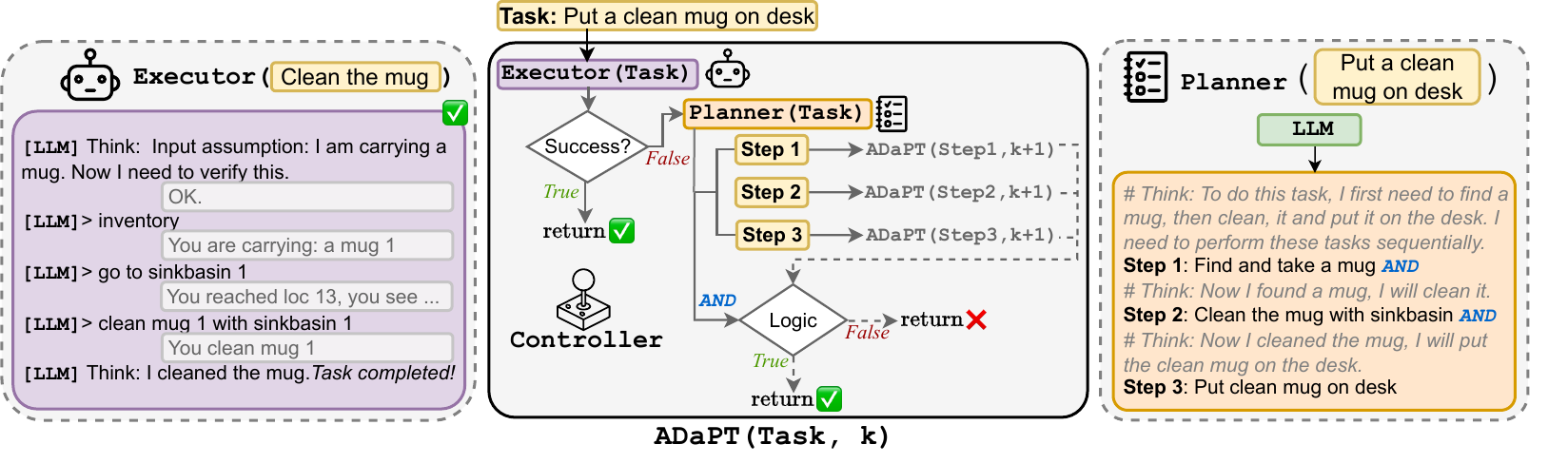}
    \caption{
    Block diagram of the \method{} pipeline with an example from ALFWorld. \textbf{Left:} Use of LLM as an executor to interact iteratively with the environment along with an example execution trajectory.  \textbf{Middle:} Overall recursive algorithm (depth $k \leq d_{\mathrm{max}}$) that embeds the executor and planner, refer to \cref{alg:main} for details. \textbf{Right:} Outline of using LLM as a planner to generate sub-tasks (steps) and logical operators combining them.
    }
    \label{fig:main}
\end{figure*}
\subsection{LLM as an \texttt{Executor} \texorpdfstring{\execfig }{}}
\label{ssec:exec}

\paragraph{Overview.} 
In a given environment, the executor is provided with a concise natural language task specification, as shown in \cref{fig:main} (left). Following \citet{yao2023react}, the executor iteratively interacts with the environment via actions generated by the LLM. This interaction continues until the task is either completed or a preset maximum iteration limit is reached.
Consistent with \citet{ahn2022can}, we provide the LLM with in-context demonstrations of low-level ``atomic'' skills specific to the environment (listed in \cref{tab:atomic} of \cref{app:details}), such as knowing how to correctly heat objects in ALFWorld. This approach offers two advantages: (i) it allows us to employ the same executor with environment-specific knowledge for all baselines (\cref{ssec:baselines}); and (ii) it enables the planner (discussed in \cref{ssec:plan}) to work at a higher level of abstraction, leveraging the LLM's general world knowledge.

\paragraph{Execution Capabilities of an LLM.} At a minimum, the LLM executor should reliably execute atomic skills. While we provide demonstrations for successful execution of atomic skills, LLMs can adapt to failures by combining multiple skills to perform complex tasks, as discussed in \cref{ssec:rq2}. For instance, in \cref{fig:main} (left), we show the LLM successfully cleaning a mug it's carrying (an atomic skill). An advanced executor could  combine ``finding a mug'' with the ``cleaning'' skill to accomplish ``find a clean mug'' without an explicit planner.

\paragraph{Self-generated Success Heuristic.} 
In order to decompose based on the abilities of the executor, we need to determine whether the executor is capable of finishing the given (sub-)task independently or if further decomposition is required. To this end, we employ the executor LLM to determine the completion of the (sub-)task \emph{without relying on the environment} for obtaining gold rewards for (sub-)tasks.
We include a simple instruction in the executor prompt to output \emph{``task completed''} if it determines it has succeeded, otherwise output \emph{``task failed''} in case it cannot proceed. Refer to example in \cref{fig:main} (left). Our success heuristic aligns with binary classification models employed in \citet{shinn2023reflexion}, providing a way to simulate intermediate rewards, which complements end-of-task environment rewards~\cite{rengarajan2022reinforcement}. We study this LLM-generated heuristic in \cref{ssec:comm} and show that it closely matches the gold reward.

\subsection{LLM as a \texttt{Planner} \texorpdfstring{\planfig}{}}
\label{ssec:plan}

\paragraph{Overview.} The objective of the planner is to break down complex tasks into smaller sub-tasks.
To achieve this, we instruct the LLM to generate a concise yet comprehensive plan consisting of a few steps, typically 3-5, as shown in \cref{fig:main} (right). We opt for shorter, more abstract plans because expecting a detailed, fine-grained plan upfront can be impractical, especially in unexplored environments. E.g., devising a 10-step plan to put a clean mug on a desk without prior knowledge of the mug's location can lead to cascading errors due to incorrect assumptions. Therefore, we task the LLM to generate short plans, with the \emph{flexibility to decompose further} in subsequent iterations, based on the executor's capabilities. 

\paragraph{Composition Logic for Sub-tasks.} Along with the sub-tasks, we prompt the planner to generate logical operators to combine various sub-tasks in the plan to accomplish the task. We allow for two logical operators: ``\textsc{And}'' and ``\textsc{Or}''.  Sub-tasks are linked using \textsc{And} when they must be executed sequentially for the task to succeed.  However, in cases requiring exploration, such as finding an item in an unknown room, we employ the \textsc{Or} operator to simulate conditional checks. Here, the task succeeds if any of the sub-tasks are successful. For instance, in \cref{fig:intro}, the plan to \emph{``find a mug''} would be to \emph{``find a mug on the countertop'' \textsc{Or} ``find a mug in the cabinet''}.  We execute the latter only if the agent has not found the mug yet. While examples in \cref{fig:intro,fig:main} show homogeneous logic, \method{} can handle complex logical expressions as described in \cref{app:logic}.

\subsection{\texttt{Controller}  -- LLM Program \texorpdfstring{\controlfig}{}}
\label{ssec:cont}

\paragraph{Overall Pipeline.} Thus far, we describe two LLM-based modules that can perform the roles of low-level execution and high-level planning.
We incorporate these modules into \method{} via the controller which is a pre-determined and recursive algorithm -- making the overall pipeline of \method{} an LLM program~\cite{schlag2023large, dohan2022language}, shown in \cref{alg:main}. The overall flow of the controller program is as follows: (i) given an input task, the controller calls the executor to check if it can succeed in performing the task directly; (ii) if the executor does not succeed, the controller delegates decomposing the complex task to the planner and recursively calls \method{} for each sub-task until we hit a termination criterion, i.e., if a maximum depth $d_{\mathrm{max}}$ ($\geq \!\! 1$) is reached.

\cref{fig:main} (mid) shows the control flow of \method{}.  A complex task such as ``put a clean mug on the desk'' is first assigned to the executor. If the executor does not succeed, then \method{} calls the planner to decompose the task into sub-tasks along with a logical operator (\textsc{And} or \textsc{Or}) indicating how to compose them. Each sub-task (referred to as `step' in \cref{fig:main}) is then assigned recursively to \method{} and is combined using the logical operator. In the end, the success of sub-tasks 
after recursive decomposition ensures overall task success (unrolled calls to planner and executor are shown in \cref{fig:intro}).

\section{Experimental Setup}
\label{sec:setup}
We describe the datasets used in our experiments and baselines used for comparison with \method{}.

\subsection{Datasets}
\label{ssec:data}
We employ LLMs-as-agents to perform tasks in the following three environments and use task \textbf{success rate} as our evaluation metric in \cref{sec:result,sec:disc}.
\paragraph{ALFWorld.}
ALFWorld~\cite{sridhar2021alfworld} is a text-based game version of the embodied ALFRED benchmark~\cite{shridhar2020alfred} implemented in the TextWorld environment~\cite{cote2019textworld}.
It encompasses 6 distinct task types, where an agent is required to accomplish high-level tasks through navigation and interaction via text-based actions in a simulated household that gives textual feedback to an agent (e.g., \emph{put a clean mug on desk} discussed earlier in \cref{fig:main}).
Following \citet{sridhar2021alfworld}, we present results on 134 unseen evaluation games (test set) with a separate dev set of 10 games per task from the seen evaluation games split. Along with atomic skills, we add example gold trajectories, following \citet{yao2023react}, for two tasks: heat and look in the executor prompt.\footnote{
Unlike \citet{yao2023react}, we use a standardized executor prompt for all ALFWorld tasks, avoiding the agent to know the task-type apriori. \cref{tab:alt_react} in \cref{app:react} further demonstrates that \method{} still improves over task-specific executors. \label{foot:hybrid}
}

\paragraph{WebShop.}
WebShop~\cite{yao2022webshop} is an online shopping website environment featuring 1.18 million real-world products containing 500 user queries in the test set.
It serves as a complex decision-making environment with practical applications wherein an agent must navigate a website through a variety of commands to purchase an item matching a user specification (e.g., \emph{grey sectional sofa priced less than \$300 with fast delivery}). Following~\citet{shinn2023reflexion}, we report performance on 100 user instructions and use a different subset of 40 queries as the dev set. 

\paragraph{TextCraft.}  We create a new text-only environment for crafting Minecraft\footnote{\url{https://www.minecraft.net}} items similar to WordCraft~\cite{coenen2021wordcraft}. Unlike existing agent-based environments, tasks in TextCraft exhibit a natural compositional structure, resembling cooking recipes with steps of varying complexity, where some sub-tasks are more intricate, such as layering a lasagna, while others are simpler, like baking it.

\begin{figure}[t]
    \centering
    \includegraphics[scale=0.72]{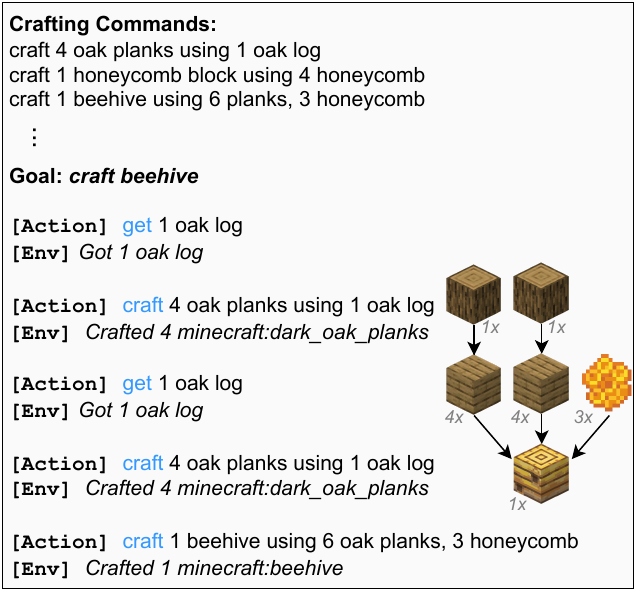}
\caption{Example gold trajectory in TextCraft for a task with recipe depth of 2.}
\label{fig:textcraft}
\end{figure}

\begin{table*}[t]
\parbox{.575\linewidth}{
    \small
    \centering
    \setlength{\tabcolsep}{2.5pt}
    \begin{tabular}{l c c c c c c | c}
    \toprule
      \bf Method  ($d_{\mathrm{max}}=3$) & \bf Pick & \bf Clean & \bf Heat & \bf Cool & \bf Look & \bf Pick2 & \bf All \\
      \midrule
    ReAct & 33.3	& \underline{67.7} &	43.5 &	33.3 &	55.6 &	\underline{11.8} &	43.3\\
    Plan-and-Execute & 29.2 &	61.3 &	47.8 &	38.1 &\bf{61.1} & 	\underline{11.8} &	43.3 \\
    Try Again with ReAct & 50.0 &	51.6 &	\underline{60.8}&	47.6 &	\bf{61.1} &	\text{ } 5.9 &	47.8\\
    Reflexion & 70.8 &	61.3 &	\bf 61.0 & 	\underline{66.7} &	\bf 61.1 &	\text{ } 5.9 &	\underline{57.5}\\
    \method{} (Ours) & \bf 87.5 &	\bf 80.6	& \underline{60.8} & 	\bf{76.2} &	\bf{61.1} &	\bf 52.9 &	\bf 71.6\\
    \bottomrule
    \end{tabular}
    \caption{
    \method{} yields the highest the overall success rates (\%) compared to baselines from prior work (discussed in \cref{ssec:baselines}) on ALFWorld (test split). Best (highest) success rates are highlighted in bold and second-highest rates are underlined. 
    }
    \vspace{-1em}
    \label{tab:alf}
    }
    \hfill
\parbox{.4\linewidth}{
    \small
    \centering
    \setlength{\tabcolsep}{2pt}
    \begin{tabular}{l c c}
    \toprule
      \bf Method  & \bf WebShop & \bf  TextCraft\\
      \midrule
    ReAct  & 32.0 \text{ } & 19.0\\
    Plan-and-Execute  & 17.0 \text{ }& 27.0\\
    Try Again with ReAct & 30.0 \text{ }& 15.0\\
    Reflexion &  35.0$^\dagger$ & \underline{32.0}\\
    LATS~\cite{zhou2023language} &  \underline{38.0}$^\dagger$ & $-$ \\
    \method{} (Ours) & \bf 44.0 \text{ } & \bf 52.0\\
    \bottomrule
    \end{tabular}
    \caption{\method{} yields the highest success rate on WebShop and TextCraft (test split) with $d_{\mathrm{max}}=3$ and $4$ respectively. $^\dagger$Performance reported by \citet{zhou2023language}}
    \label{tab:web-text}
    \vspace{-1em}
    }
\end{table*}

Tasks in TextCraft are inherently decomposable. In \cref{fig:textcraft}, crafting a beehive necessitates crafting its ingredients, like planks and honeycomb, which may require further decomposition. The agent thus needs to identify and adapt to varying task complexity, e.g., crafting a plank is \emph{easier} than crafting a beehive. Moreover, some recipes allow using any item from a particular category. For instance,  crafting a beehive uses planks (a category), requiring the agent to use linguistic knowledge for proper item selection (e.g., select oak planks, a specific item in the category planks).
We evaluate our approach on a test set of 200 tasks where the target items have recipe trees of depth 2, 3, and 4 (example tree of depth 2 is shown in \cref{fig:textcraft}). We use the items with recipe tree depth of 3 (123 tasks), depth of 4 (11 tasks) and depth of 2 (77 out of 297) in our test set, and the rest of depth 2 tasks constitute the dev set. Additional details about creating the environment are present in \cref{app:textcraft}.

\subsection{Baseline Approaches}
\label{ssec:baselines}

We compare \method{} with four classes of baseline approaches described below.

\paragraph{Iterative Executor-Only (ReAct).} In this setting, we employ the executor to interact iteratively with the environment, adopting the think-act-observe prompting style from ReAct~\cite{yao2023react}. All methods discussed below, including \method{}, share the \emph{same} executor, ensuring a standardized impact of the executor's strength and design choices when comparing relative performance in \cref{sec:result}. When $d_{\mathrm{max}}\!=\!1$, \method{} solely relies on this executor.

\paragraph{Plan-and-Execute.} As shown in \cref{fig:intro}, in this setting, we generate a plan first and then assign each sub-task to the executor. This approach only plans once and as a result has a non-adaptive structure (consistent with \citet{wang-etal-2023-plan, yang2023intercode,Sun2023PEARLPL}). To ensure each plan step is executable without further decomposition, we design new prompts with more detailed plans.
Note that \method{} with $d_{\mathrm{max}}\!=\!2$ differs from plan-and-execute as it is adaptive, i.e., decomposes only when executor fails and generates relatively shorter plans (refer to \cref{app:logic}).

\paragraph{Try Again with ReAct.} By design, \method{} makes multiple calls to the executor module, albeit with different (sub-)tasks.   
Like~\citet{yang2023intercode}, we design a simple controller that requests the executor to retry the task in a total of $d_{\mathrm{max}}$ separate trials and then uses the trial with the best performance for each task instance.

\paragraph{Reflexion.} \citet{shinn2023reflexion} execute the entire task first, and if unsuccessful, reflect and store feedback in memory for subsequent $d_{\mathrm{max}} \! - \! 1$ trials. While adaptive, this approach repeats the entire trial even if a single sub-task fails, redundantly re-executing previously successful sub-tasks.

\paragraph{\method{} and Shared Implementation Details.} Following \cite{yao2023react, shinn2023reflexion, zhou2023language}, by default, we use the GPT-3.5~\cite{ouyang2022training} LLM for both planning and execution in \method{} and other baselines. We use the completion-based models for ALFWorld and TextCraft and the chat-based model for WebShop.\footnote{We use the completion model as chat variants of GPT-3.5 consistently underperform their completion counterparts~\cite{liu2023agentbench, yang2023intercode}. We discuss the effectiveness of \method{} different LLMs in \cref{ssec:rq2}.\label{foot:model}} Further, we use \method{} (and other baselines) with $d_{\mathrm{max}}\!=\!3$ for ALFWorld, and WebShop and increase to $d_{\mathrm{max}}\!=\!4$ for TextCraft to accommodate recipes with a depth of 4 (\cref{ssec:data}). For additional details, refer to \cref{app:details}. We increase the maximum number of iterations for the ReAct baseline by a factor of $d_{\mathrm{max}}$ and ensure all baselines use a comparable number of LLM calls (\cref{app:calls}).

\section{Main Results}
\label{sec:result}
Using GPT-3.5 as the underlying LLM, in this section, we show that \method{} yields the highest success rate compared to baselines from prior work on ALFWorld, WebShop, and TextCraft datasets.

\paragraph{ALFWorld.} In \cref{tab:alf}, we observe that \method{} achieves the \emph{highest overall success rate}, while using  ReAct alone results in the lowest overall performance. 
By leveraging adaptive decomposition, \method{} improves over ReAct's performance by $28.3\%$ points (absolute) as well as over Plan-and-Execute and Try Again by $28.3\%$ and $23.8\%$ points, respectively. Lastly, we find that \method{} yields $14.1\%$ points higher overall success rate than Reflexion, despite the latter having access to dedicated memory and natural language feedback. 
Specifically, we find baselines yield poor results on `pick2' tasks ($<\! \! 12\%$ success rate) as they require the agent to compose two `pick'-style tasks involving a longer action history. However, \method{} yields significant improvements (by over a factor of $4\times$) for this type of tasks.

\begin{figure}[t]
    \centering
    \includegraphics[trim={0cm 0cm 0 0},clip,scale=0.5]{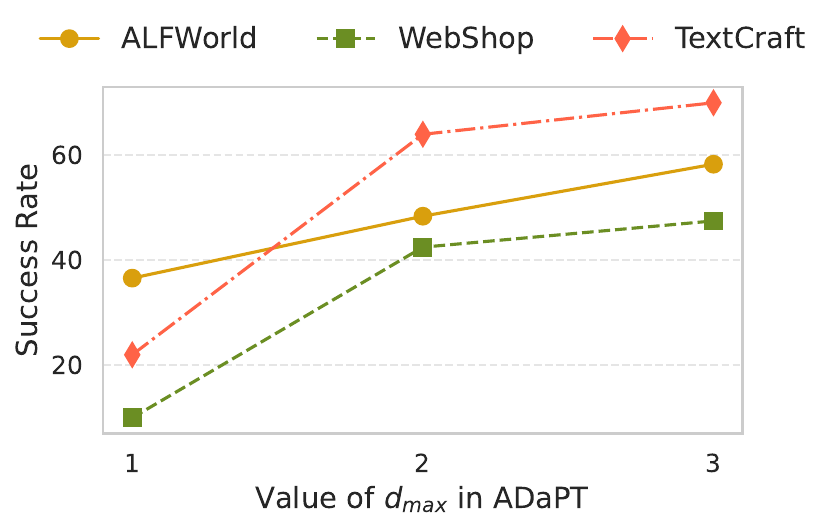}
    \caption{Success rate of \method{} increases with the maximum depth $d_{\mathrm{max}}$ for all datasets (dev splits).}
    \label{tab:depth}
\end{figure}

\paragraph{WebShop.} \cref{tab:web-text} shows a similar trend with \emph{\method{} surpassing all baselines} and achieving the highest success rate. \method{} outperforms ReAct, Plan-and-Execute, and Try-Again baselines by up to $27\%$ points. We corroborate the findings of \citet{shinn2023reflexion} and observe that natural language feedback offers limited gains in performance, as compared to \method{} (which surpasses Reflexion by $9\%$ points). 
Additionally, we compare with a recent search-based baseline LATS~\cite{zhou2023language} and find that \method{} outperforms the success rate of LATS by $6\%$ points. 

\paragraph{TextCraft.} Our results on TextCraft are summarized in \cref{tab:web-text}. First, we observe that \method{} \emph{achieves an improvement of $33\%$} compared to the ReAct executor. In contrast to Plan-and-Execute, i.e., starting with a fixed plan, having the dynamic ability to adapt to complex sub-tasks (in this case, crafting complex ingredients) in \method{} improves performance by $25\%$ points. Lastly, \method{} outperforms Reflexion by $20\%$ points, highlighting the importance of adaptive and as-needed planning.
\edited{
We hypothesize that \method{} consistently outperforms Reflexion across datasets as the latter relies on generating feedback based on errors in the entire trajectory. In contrast, due its design, \method{} often handle failures of small sub-tasks and redirects more resources in the form of calling the planner and decomposition to the challenging sub-tasks.
}

\section{Analysis and Discussion}
\label{sec:disc}

We analyze \method{} in detail by addressing the following research questions on dev data splits.

\subsection{How does performance of \method{} scale with the depth of decomposition?}
\label{ssec:rq1}
\paragraph{Setup.} To assess the impact of adaptive decomposition, we study \method{} under three settings with increasing maximum depth $d_{\mathrm{max}} \in \{1, 2, 3\}$ for ALFWorld, WebShop, and TextCraft. Note that $d_{\mathrm{max}}\!=\!1$ setting corresponds to the iterative executor-only baseline (ReAct).

\paragraph{Results.} 
\cref{tab:depth} shows that across all datasets, performance of \method{} scales with increasing the maximum depth $d_{\mathrm{max}}$.
Consistently, we find a significant improvement in success rates as we move from $d_{\mathrm{max}}\!=\!1$ to $d_{\mathrm{max}}\!=\!2$, i.e., adding the planner to decompose a complex task when executor fails proves to be effective. Finally, the performance increase from $d_{\mathrm{max}}\!=\!2$ to $d_{\mathrm{max}}\!=\!3$ validates our hypothesis that some sub-tasks are difficult for the LLM to directly execute successfully, and decomposing these further  boosts overall performance.

\subsection{Does \method{} cater to different execution capabilities of LLMs?}
\label{ssec:rq2}

\begin{figure}[t]
    \centering
    \includegraphics[trim={0.2cm 0cm 0 0},clip,scale=0.6]{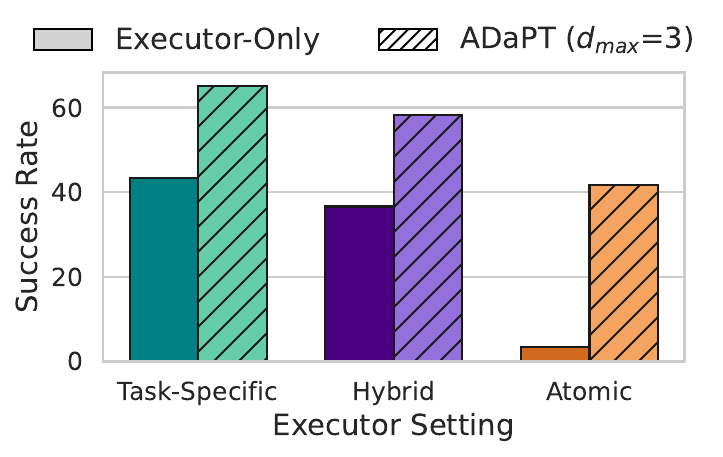}
    \caption{\method{} improves success rates across varying settings capturing different executor capabilities (i.e., executor-only performance) on ALFWorld (dev).}
    \label{fig:setting}
\end{figure}
\begin{figure*}[t]
    \centering
    \includegraphics[scale=0.6]{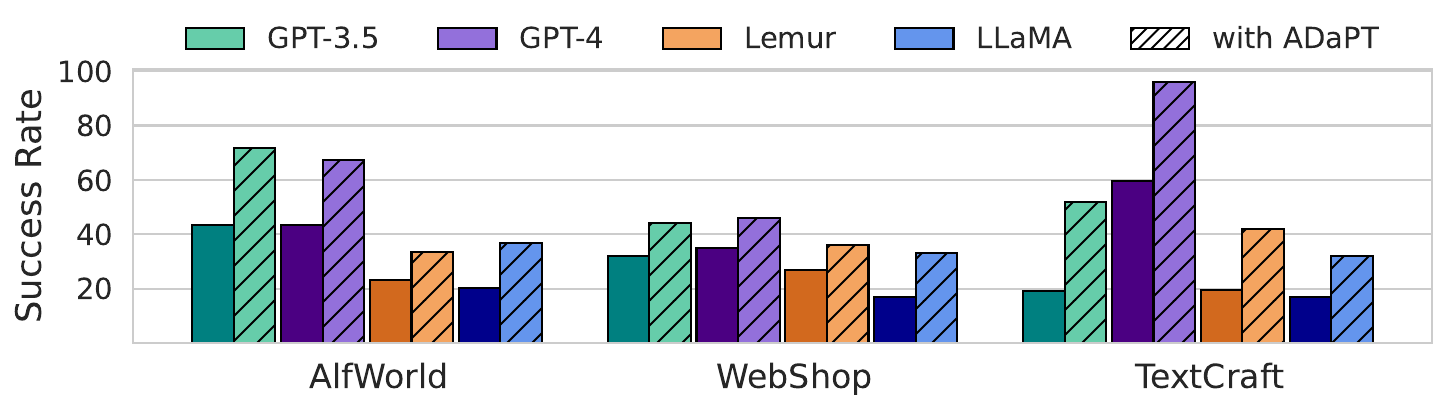}
    \caption{\method{} improves (test) performance of GPT-3.5, GPT-4, LLaMA, and Lemur LLMs  across datasets.}
    \label{fig:llm}
\end{figure*}
\paragraph{Same LLM, different execution capabilities.}
We run \method{} on three different executor prompts on ALFWorld: (i) task-specific gold trajectories, (ii) atomic skills and common gold-trajectories for 2 tasks used in \cref{sec:result} (hybrid),
and (iii) only atomic skills. Using gold trajectories aligns closely with the task at inference-time and thus, should exhibit high performance. In contrast, executor using only atomic skills relies on the inherent composition abilities of the LLM, yielding weaker performance. Here we examine if \method{} can improve success rates for all three settings.

\paragraph{Results.} In \cref{fig:setting}, we observe that \method{} consistently improves over the executor-only baseline for \emph{all diverse executor settings}. As expected, the executor prompted with task-specific trajectories performs the best (left), while the executor with only atomic skills performs the worst (right). Notably, \method{} substantially improves performance of the relatively weak executor, improving success rate from $3.3\%$ to $41.7\%$.

\paragraph{\method{} with different LLMs.} We study the ability of \method{} to improve performance across different LLMs (as planners and executors): (i) GPT-3.5, (ii) GPT-4~\cite{openai2023gpt4}, (iii) LLaMA-2 70B~\cite{touvron2023llama}, and (iv) Lemur 70B~\cite{xu2023lemur} on test splits of all datasets.
\paragraph{Results.} \cref{fig:llm} shows that \method{} consistently improves downstream performance for \emph{all} models across \emph{all} three datasets. Consistent with \citet{liu2023agentbench}, we find that the gated GPT models outperform the open-source models based on absolute success rates. Nevertheless, \method{} is effective across LLMs and improves performance of GPT-4, the strongest LLM, by up to $37\%$, as well as LLaMA, the least performant LLM, by up to $15\%$ on the TextCraft dataset.

\subsection{Does \method{} handle task complexity?}
\label{ssec:rq3}

\paragraph{Setup.}
By the compositional design of TextCraft, complexity of each task in the dataset can be defined with respect to the depth of the crafting recipe, i.e., recipes with higher depth would be more complex to craft. We evaluate efficacy of \method{} and the ReAct baseline on the test set of TextCraft with increasing recipe depth.\footnote{As we have only 11 tasks with recipe depth of 4, we exclude them from this analysis.} Furthermore, while we provide \method{} with a maximum budget of $d_{\mathrm{max}}=4$, we study how the maximum decomposition depth utilized by \method{} to succeed ($k_{\mathrm{max}}$) varies with task complexity.

\begin{table}[t]
\small
\centering
\setlength{\tabcolsep}{2pt}
\begin{tabular}{l c c c }
    \toprule
      \bf Method  & \bf Recipe Depth & \bf$\boldsymbol{k_{\mathrm{max}}}$ & \bf Success Rate \\
      \midrule
    ReAct & 2&  1.0 & 26.9\\
    \method{} ($d_{\mathrm{max}}=4$) & 2&  1.9 &  \bf 78.2\\
    \midrule
    ReAct &3 & 1.0 & \text{ } 1.8\\
    \method{} ($d_{\mathrm{max}}=4$) & 3 & 2.8  & \bf 38.7\\
    \bottomrule
    \end{tabular}
    \caption{\method{} improves TextCraft (test) performance even as recipe depth increases. The maximum decomposition depth used by \method{} to succeed at the task ($k_{\mathrm{max}}$) also scales with the recipe depth.}
    \label{tab:task2}
\end{table}

\paragraph{Results.} 
In \cref{tab:task2} we observe that \method{} improves success rates for games with recipe depth of 2 from $26.9\%$ to $78.2\%$, and of depth 3 from $1.8\%$ to $38.7\%$ as compared to the ReAct baseline. As expected, the executor alone is unable to handle complex recipes with depth $\geq 3$, but with the help of \method{} the performance improves significantly. Additionally, given the same budget $d_{\mathrm{max}} \! =\! 4$, as the recipe depth (complexity) increases from $2$ to $3$, \method{}'s level of decomposition ($k_{\mathrm{max}}$) also increases from $1.9$ to $2.8$.
This showcases that \method{} leverages as-needed decomposition in order to handle task complexity.

\subsection{Can we use different planner and executor LLMs within \method{}?}
\label{ssec:rq4}
\paragraph{Setup.}
\edited{
The planner and executor modules of \method{} do not need to necessarily use the same underlying model. Following, \citet{lin2023swiftsage} we explore if a relatively smaller LLM can be used to perform local actions in the executor and a more advanced LLM be used to devise plans. To this end, we explore different combinations of planner and executor LLM, with the latter using both gated and open-source models on ALFWorld. 
}

\paragraph{Results.}
\edited{ 
\cref{tab:combo} shows that \method{} can successfully be used to generate plans from one LLM that are useful to a different, possibly smaller, executor LLM, improving success rates by up to $19.9\%$ compared to the executor-only (ReAct) setting. Interestingly, using an open-source model, such as LLaMA-2-70B-chat~\cite{touvron2023llama} can be used as an executor with a more advanced LLMs such as GPT-3.5 to improve success rates by $22.9\%$ points. Since the planner LLM is used sparingly, open-source executors can dramatically decrease the monetary or computational costs of using \method{}. 
We defer combining knowledge from stronger and weaker LMs within \method{} to future work, as examined in the context of mathematical reasoning~\cite{fu2023specializing, saha2023can}.
}

\begin{table}[t]
\small
\centering
\setlength{\tabcolsep}{4pt}
\begin{tabular}{l c c  }
    \toprule
      \bf Executor LM  & \bf Planner LM &  \bf Success Rate \\
      \midrule
      GPT-3.5 & $-$ & 38.4\\
      GPT-3.5 &  GPT-3.5 & \bf 58.3 \\
    \midrule
     LLaMA-2-70B & $-$ & 20.4\\
     LLaMA-2-70B & GPT-3.5 & \bf{43.3}\\
    \bottomrule
    \end{tabular}
    \caption{\method{} improves performance on ALFWorld (dev) when using different planner and executor LLMs.}
    \label{tab:combo}
\end{table}

\subsection{How does \method{} compare to baselines in terms of LLM calls?}
\label{app:calls}
\paragraph{Setup.} 
\edited{
}
\edited{Performance of decision-making agents can be enhanced by increasing the number of calls allowed to an LLM, e.g., number of retrials in Reflexion. To verify that the gains in \method{} are not simply due to higher number of LLM calls, we compare the average of number of LLM calls made by \method{} to the baselines.}
\begin{figure}[ht]
    \centering
    \includegraphics[trim={0 0.2cm 0 0.25cm},clip,scale=0.6]{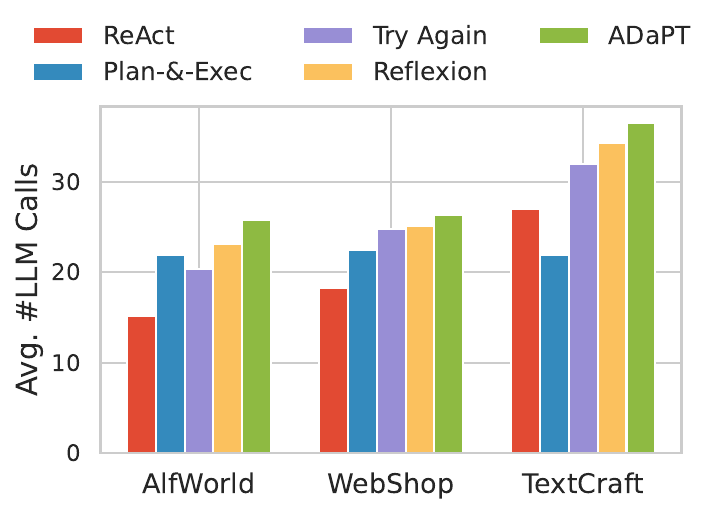}
    \caption{Average number of LLM calls for each approach including \method{} and baselines discussed in \cref{ssec:baselines} with GPT-3.5 LLM across datasets.}
    \label{fig:calls}
\end{figure}

\paragraph{Results.}
\edited{
\cref{fig:calls} shows that a \method{} employs a comparable number of LLM calls w.r.t. Try-Again and Reflexion baselines in order to yield performance improvements discussed in \cref{sec:result} (\cref{tab:alf,tab:web-text}). Note that while all methods including ReAct and Plan-and-Execute baselines are offered a comparable computational budget, the actual number of LLM calls used by the latter is often lower due to their inability to handle intermediate execution failures. This strengthens the argument for effectiveness of \method{} as the improvements do not simply stem from using substantially higher number of calls to the LLM.
}

\section{Conclusion}
We introduce \method{}, a recursive algorithm designed to harness the planning capabilities of LLMs, dynamically decomposing complex tasks when the LLM acting as an executor encounters challenges. Our evaluation across three diverse decision-making tasks, ALFWorld, WebShop, and TextCraft, reveals impressive performance of \method{}, surpassing existing baselines by substantial margins of up to $28.3\%$, $27\%$, and $33\%$ points, respectively. This not only underscores the effectiveness of \method{} but also highlights the significance of as-needed decomposition in enhancing task performance. Moreover, our findings demonstrate that \method{} not only adapts to the capabilities of the underlying executor LLM but also takes into account the complexity of individual task instances, showcasing its versatility and effectiveness.
\label{sec:conc}

\section*{Acknowledgements}
Part of this work was done during internship at AI2 and was partially supported at UNC by NSF-CAREER Award 1846185, NSF-AI Engage Institute DRL-2112635, DARPA Machine Commonsense (MCS) Grant N66001-19-2-4031,. We sincerely thank Bodhisattwa Prasad Majumder, Chris Callison-Burch, Shashank Gupta, Peter Jansen, Bill Yuchen Lin and the Aristo team for their valuable feedback. We also thank Swarnadeep Saha, Elias Stengel-Eskin, and Peter Hase for their feedback.

\section*{Limitations}
\method{} relies on the success heuristic generated by the executor LLM to determine if the model is capable of performing a complex task. For decision-making tasks studied in this work, we find that LLMs can reliably determine task success based on past action trajectories and textual feedback from the environment (see \cref{ssec:comm}). However, \citet{huang2023large, stechly2023gpt4} discuss the limits of LLM's ability to self-evaluate and self-refine. In such situations, future works may additionally employ external verifiers~\cite{lightman2023let, shridhar2023art}, theory-of-mind strategies among multiple LMs~\cite{saha2023can}, and other calibration and self-evaluation techniques~\cite{kadavath2022language}. These improved self-evaluation techniques could be useful to extend our framework to non-decision making tasks such as question answering. 
\bibliography{anthology,custom}

\appendix
\label{sec:appendix}
\section{\method{} Implementation Details}
\label{app:details}
\begin{table}[t]
    \small
    \centering
    \setlength{\tabcolsep}{3pt}
    \begin{tabular}{c c p{0.6\linewidth}}
    \toprule
         & \bf Atomic Skill &  \multicolumn{1}{c}{\bf Description}  \\
    \midrule
    \multirow{8}{*}{\rotatebox[origin=c]{90}{\textbf{ALFWorld}}} & put  & Assuming that the robot is carrying an object, put it on a given receptacle.\\
    & take & Take a specified object from a specified receptacle.\\
    & clean/heat/cool & Assuming that the robot is carrying an object, clean/heat/cool the object.\\
    & examine & Assuming the robot is at a desk with a desk lamp, use it to look at an object.\\
    \midrule
    \multirow{9}{*}{\rotatebox[origin=c]{90}{\textbf{WebShop}}} & search  & Put a given query in the search box, results in a page with list of products.\\
    & shortlist & Based on the search page and query, get list of any matching products.\\
    & match & Given a product ID and query, navigate to the product page and verify it matches the query.\\
    & buy & Given a product ID and query, buy product by selecting relevant options.\\
    \midrule
    \multirow{7}{*}{\rotatebox[origin=c]{90}{\textbf{TextCraft}}} & craft  & Assuming the agent has all the ingredients in the inventory, craft a target object by picking an appropriate command from the list of crafting recipes.\\
    & fetch & Look for a given object in the inventory or get it directly from the game.\\
    & inventory & Look-up the game inventory.\\

    \bottomrule
    \end{tabular}
    \caption{Overview of atomic skills used in \cref{ssec:exec}.}
    \label{tab:atomic}
\end{table}

\paragraph{Executor.} We use a common ReAct executor for each dataset. To this end, we provide the LLM in the executor  with in-context example trajectories for each atomic skill (refer to \cref{tab:atomic} for an exhaustive list). Atomic skills are inherently task dependent, and thus, vary with the underlying environment. For ALFWorld, in which the agent needs to navigate and perform tasks in the household, the atomic skills include: taking an object, putting it down at a location, cleaning, heating, etc. On the other hand, the goal in WebShop is to buy a product based on user queries, thus, atomic skills include: searching a specified query, shortlisting products based on search page, matching if a product satisfies a criteria, and buying a product. Lastly, the atomic skills in TextCraft are fetching objects from the environment, and crafting them given the recipe and the ingredients. Following \citet{yao2023react}, we add gold trajectories for two tasks: heat and look in the executor prompt for ALFWorld,
and one full gold trajectory for TextCraft.

\paragraph{Planner.} We provide the LLM with a brief description of atomic skills and in-context demonstrations of few task decompositions for each dataset.
\begin{itemize}[leftmargin=*,noitemsep,nolistsep]
\item \textbf{ALFWorld:} The planner includes 6 demonstrations of task decompositions for one household configuration. Specifically, \emph{``find''} is not an atomic skill for the executor, and therefore, needs to be handled by the planner (refer to \cref{fig:main}).
\item \textbf{WebShop:} The planner breaks down a given task in terms of the atomic skills described in \cref{tab:atomic} via 2 in-context demonstrations.
\item \textbf{TextCraft:} The planner determines the necessary ingredients for each item and creates a plan to obtain them and then craft the item, illustrated via 2 examples with different crafting commands.
\end{itemize}

\begin{table*}[t]
\parbox{.55\linewidth}{
    \small
    \centering
    \setlength{\tabcolsep}{2.5pt}
    \begin{tabular}{l c c c c c c | c}
    \toprule
      \bf Method  & \bf Pick & \bf Clean & \bf Heat & \bf Cool & \bf Look & \bf Pick2 & \bf All \\
      \midrule
    ReAct & 66.7	& 41.9 & 47.8 &	80.9 &	\underline{83.3} &	23.5 &	56.7\\
    Plan-and-Execute & \underline{87.5} & 58.1 & \underline{73.9} & 52.4 & \underline{83.3} & 17.6 & 63.4\\
    Try Again with ReAct & 75.0 & 38.7 &60.9 & 76.2 & 66.7 & 23.5 & 56.7 \\
    Reflexion &	83.3 & \underline{61.3} &  \underline{73.9}& \bf 85.7& 61.1 & \underline{29.4} & \underline{67.2}\\
    \method{} (Ours) & \bf 91.7 & \bf 67.7 & \bf 78.3 & \underline{81.0} &\bf  100 & \bf 64.7 &\bf 79.8\\
    \bottomrule
    \end{tabular}
    \caption{Comparison of success rates (\%) achieved by \method{} and other baselines from prior work  on ALFWorld (test split) with executor used by \citet{yao2023react}}
    \vspace{-1em}
    \label{tab:alt_react}
    }
    \hfill
\parbox{.375\linewidth}{
    \small
    \centering
    \setlength{\tabcolsep}{2.5pt}
    \begin{tabular}{l c c}
    \toprule
      \bf Method  & \bf Score &\bf Success Rate \\
      \midrule
    Iterative Executor-Only &42.1 & 29.0\\
    Static Decomposition & 27.7 & 17.0\\
    Retry Execution & 45.4 & 30.0\\
    \rowcolor{light-gray} Naive & 58.3 & 24.0\\
    \rowcolor{light-gray} Reflexion* & 64.2 & 35.0\\
    \rowcolor{light-gray}  LATS~\cite{zhou2023language}* & 75.9 & 38.0 \\
    \method{} (Ours) & 60.0 & \bf 44.0\\
    \bottomrule
    \end{tabular}
    \caption{Performance comparison of different methods on WebShop.}
    \label{tab:web-score}
    \vspace{-1em}
    }
\end{table*}

\begin{figure*}[!ht]
    \centering
    \includegraphics[scale=0.6]{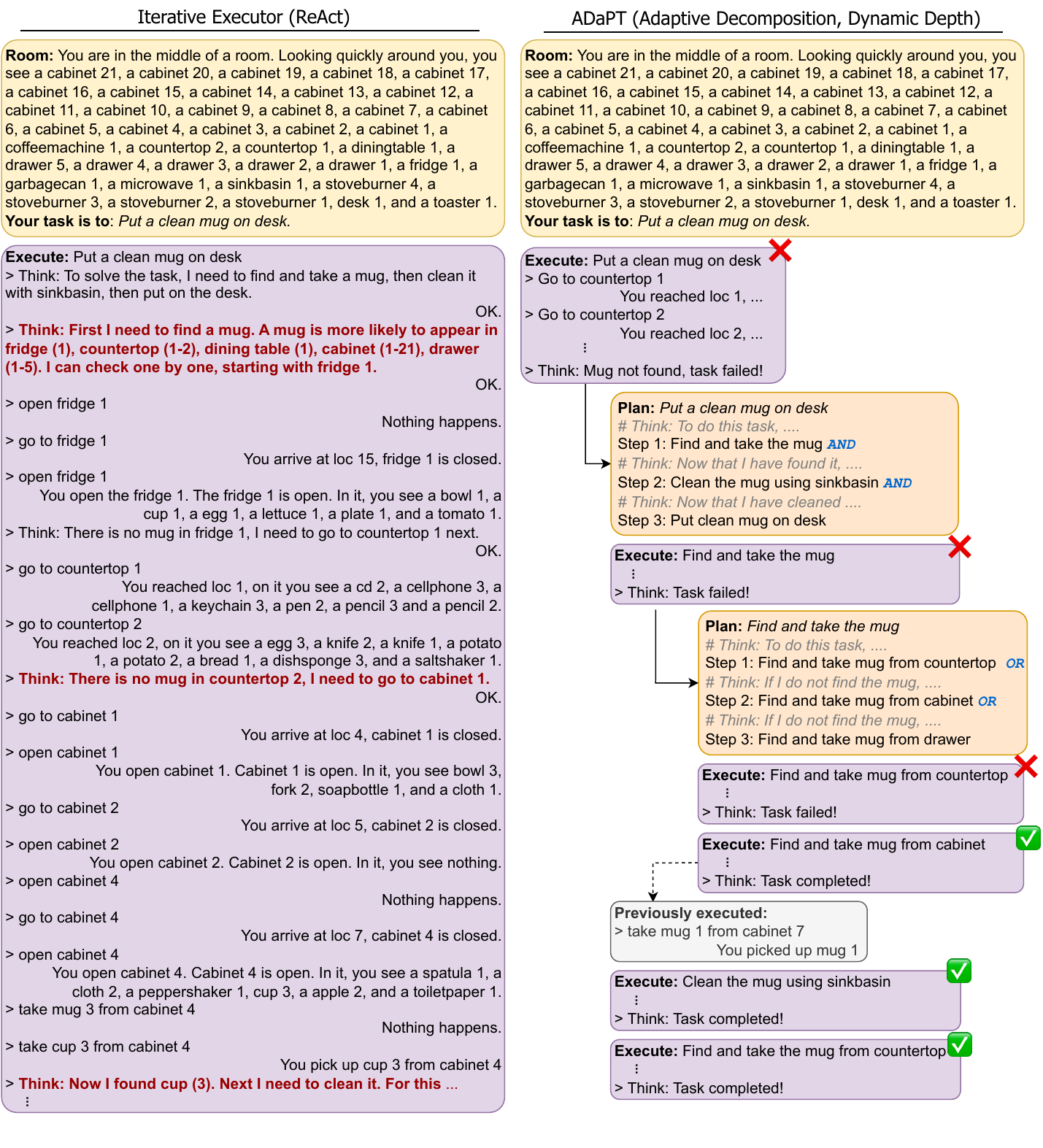}
    \caption{Comparison of iterative executors such as ReAct with \method{}. On left, ReAct uses interleaved ``thought'' statements to set milestones and track their progress. However, due to a large action history, it struggles to follow the plan exactly and hallucinates the wrong object (highlighted in red). \method{}, on the right, decomposes complex tasks into smaller sub-tasks whenever the executor fails, leading to shorter action trajectories for easy execution.}
    \label{fig:alf_roll}
\end{figure*}

\begin{figure}[!ht]
    \centering
    \includegraphics[scale=0.575]{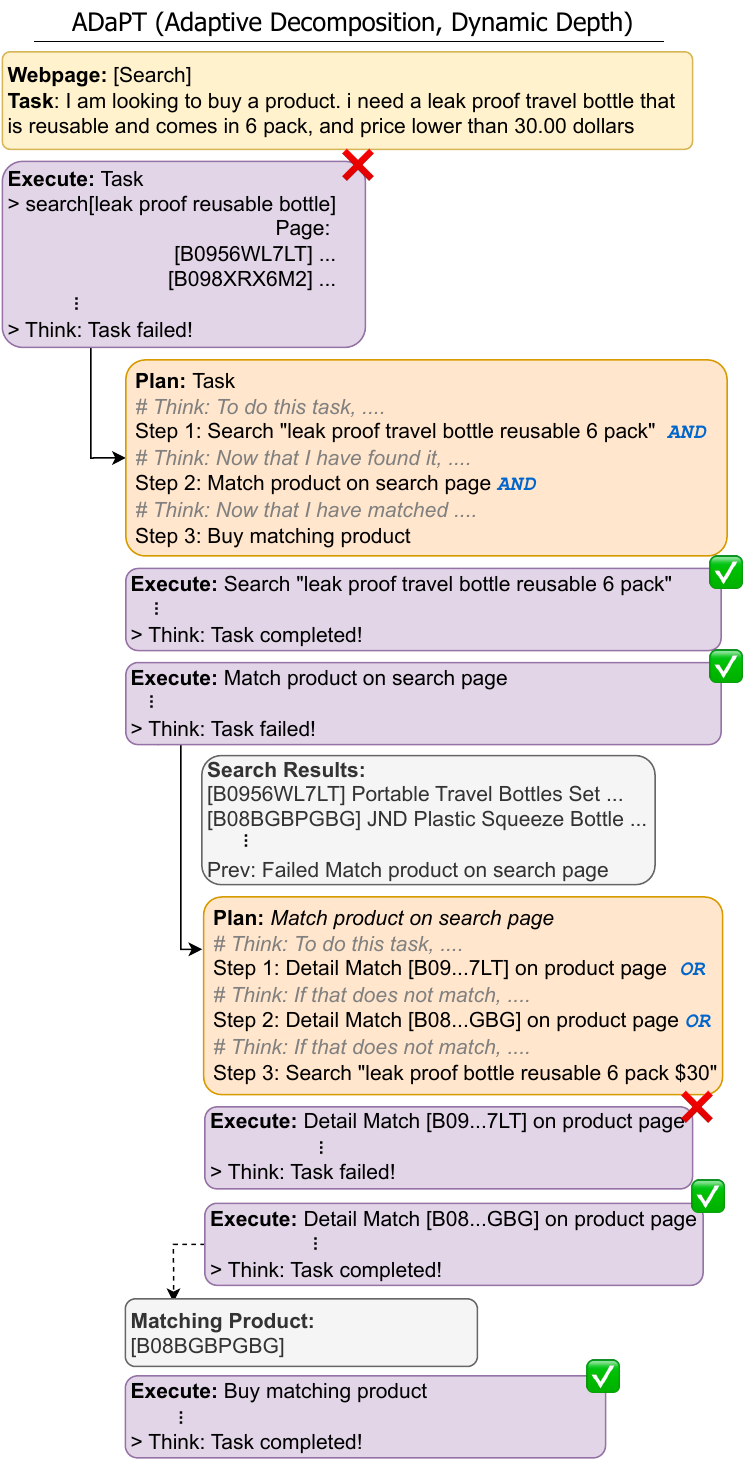}
    \caption{Partial rolled out trajectories for WebShop with \method{}. In the gray box we communicate to the planner the current (search) page that is visible to the agent, and once a matching product is found, we propagate it to future executor runs. Note ``match on search page'' corresponds to shortlist skill in \cref{tab:atomic}, and ``detail match on product page'' corresponds to match skill.}
    \label{fig:web_roll}
\end{figure}

\begin{figure}[!ht]
    \centering
    \includegraphics[scale=0.575]{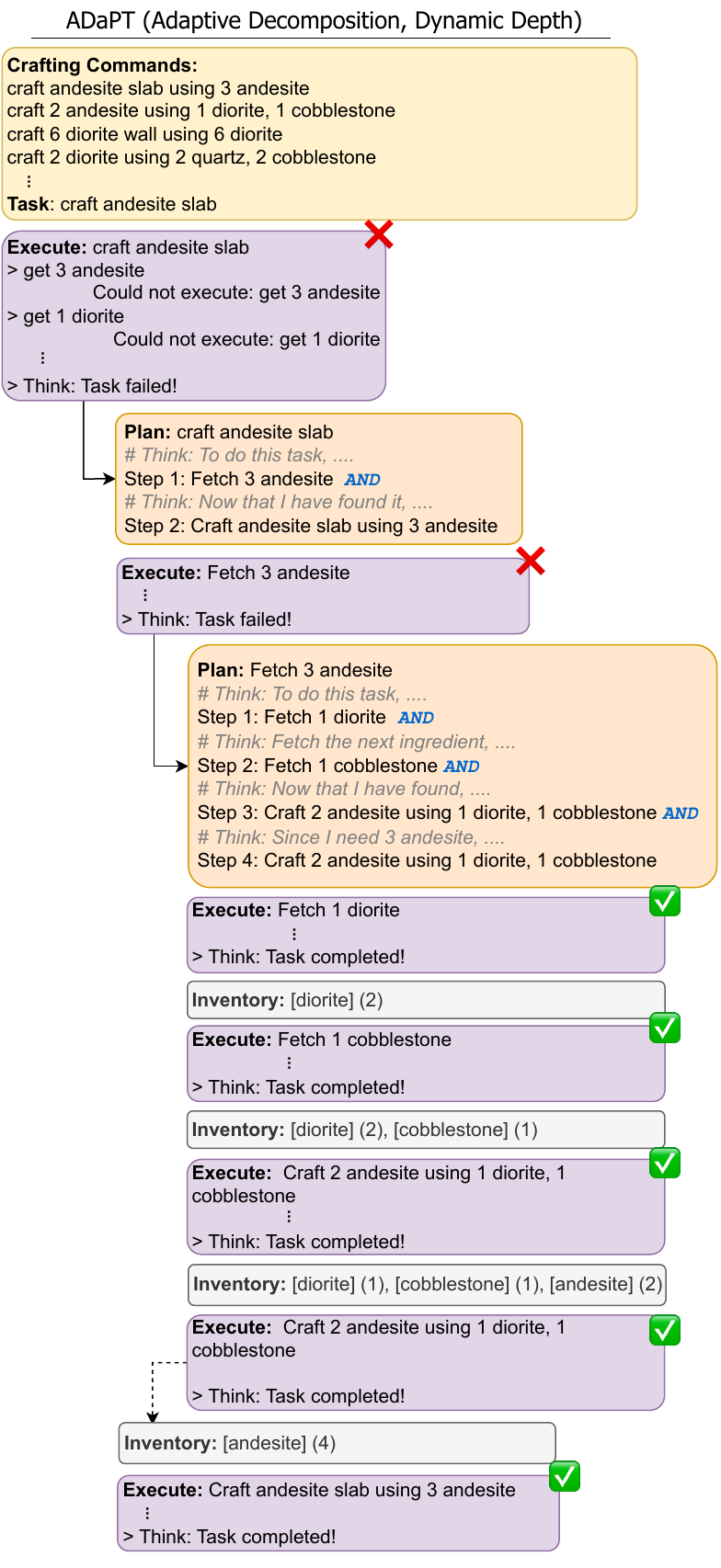}
    \caption{Partial rolled out trajectories for TextCraft using \method{}. In the gray box, we propagate the inventory of the agent to subsequent executor calls. Note that while ``diorite'' is not directly present in the environment, i.e., it needs to be crafted. The executor LLM is able to inherently compose skills to fetch it without further decomposition.}
    \label{fig:text_roll}
\end{figure}

\begin{algorithm}[t]
\begin{algorithmic}[1]
\small 
\Function{\method{}}{Task $T$, Current depth $k$}
\State \textcolor{comm}{\footnotesize{\textit{// \method{}$(\cdot)$ Generates success heuristic value $completed$ for the task $T$. Initialized with $k=1$.}} }
\State \textcolor{comm}{\footnotesize{\textit{// Base case: terminate on reaching  maximum depth}}} 
\IfThen{$ k > d_{\mathrm{max}}$}{\Return $False$}  \label{line:3}
\State \textcolor{comm}{\footnotesize{\textit{// Execute the task/sub-task to assess if the LLM can directly perform it using LLM-generated $success$.}}} 
\State $completed \gets \boldsymbol{\mathrm{executor}_{\textsc{llm}}}(T)$ \label{alg:line1}
\State \textcolor{comm}{\footnotesize{\textit{// Plan only when the executor fails.}}} 
\If{$completed \text{ is } False$}
\State \textcolor{comm}{\footnotesize{\textit{// Using the LLM, decompose the task into a set of sub-tasks, $\mathcal{P}$, and a Boolean function, $logic(\cdot)$, that combines output of the sub-tasks.}}}
\State $\mathcal{P}, logic \gets \boldsymbol{\mathrm{planner}_{\textsc{llm}}}(T)$   \label{line:5}
\State \textcolor{gray}{\footnotesize{\textit{// Get the outputs for individual sub tasks}}}
\State $\mathcal{O} = \{\textbf{\method{}}(T_{\mathrm{sub}}, k \! + \! 1)|{T_{\mathrm{sub}} \in \mathcal{P}}\}$
\State \textcolor{gray}{\footnotesize{\textit{// Combine the outputs of the sub tasks}}}
\State $completed \gets logic(\mathcal{O})$
\EndIf
\State \Return $completed$
\EndFunction

\caption{Algorithm for \method{}
}
\label{alg:main}
\end{algorithmic}
\end{algorithm} 
\paragraph{Controller.} The controller performs two crucial roles in the overall functioning of \method{}. First, it serves as the \emph{communication bridge} between planner and executor, propagating salient information across the two depending on the task. Second, since \method{} is a recursive algorithm, the controller determines the \emph{termination criterion} using the logical expression from the planner and success heuristic from the executor or if a maximum depth $d_{\mathrm{max}}$ ($\geq \!\! 1$) is reached. The controller propagates task-dependent salient information described below:
\begin{itemize}[leftmargin=*,noitemsep,nolistsep]
\item \textbf{ALFWorld:  }In the controller, we propagate the last successful action from a previous execution run to subsequent calls of the executor. Note that information is only propagated from successful sub-tasks. For sub-tasks connected via ``\textsc{Or}'', each receives the same information from the controller. Unlike \citet{shinn2023reflexion}, executor does not get text feedback from prior failures.
\item \textbf{WebShop: } We propagate the current page visible to the agent along with past unsuccessful executor tasks to the planner (without any rationales). Once we find a matching product, we also propagate the product ID in future executor calls.
\item \textbf{TextCraft: } We propagate the current inventory of the agent to the executor. This is akin to executors starting with the \texttt{inventory} command as the first step to keep stock of which items are missing and need to be fetched or crafted. 
\end{itemize}
For partial rolled-out trajectories with \method{} refer to \cref{fig:alf_roll,fig:web_roll,fig:text_roll}. Communication between planner and executor is highlighted in \colorbox{light-gray}{gray box(es)}.

\paragraph{LLM-related Hyperparameters.} Following previous works~\cite{shinn2023reflexion, liu2023agentbench} we use \texttt{text-davinci-003} from the OpenAI API for ALFWorld. For WebShop, we use the \texttt{gpt-3.5-turbo} models, and for TextCraft we use the \texttt{gpt-3.5-turbo-instruct} models. All executors have a maximum budget of iterations  to interact with the environment and execute the task. We set this budget to 20, 15, and 20 respectively for ALFWorld, WebShop, and TextCraft respectively. For try again with ReAct, we sample additional trajectories with a temperature of 0.7. As discussed in \cref{ssec:baselines}, we run the iterative executor-only baseline for 60, 45, 60 iterations for ALFWorld, WebShop, and TextCraft respectively. In \cref{ssec:rq2}, we use publicly available checkpoints for LLaMA 70B\footnote{\url{https://huggingface.co/meta-llama/Llama-2-70b-hf}} and Lemur 70B\footnote{\url{https://huggingface.co/OpenLemur/lemur-70b-chat-v1}} available on Huggingface~\cite{wolf2019huggingface}. For both planner and executor modules, we use a fixed prompt consisting of few in-context examples (as described above) for each dataset. We show all executor and planner prompts to the LLM in \cref{app:prompts}. Due to cost constraints, we report success rates for a single run of each LLM in \cref{sec:result,sec:disc}.

\begin{figure}[t]
    \centering
    \includegraphics[scale=0.7]{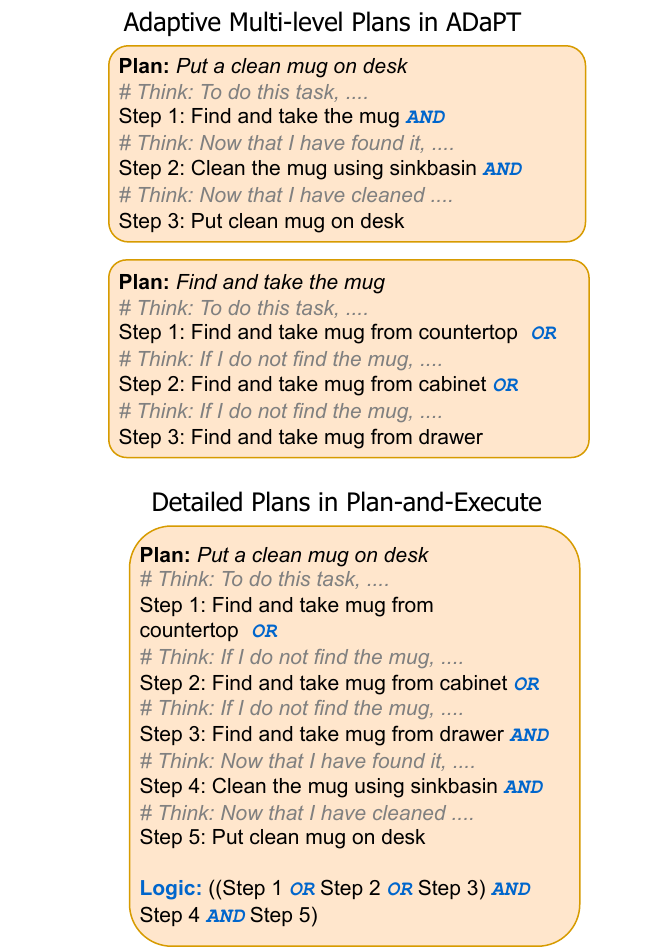}
    \caption{Illustration of how multiple levels of plans from \method{}, can be collapsed into one detailed plan in non-adaptive settings as used in the plan-and-execute baseline (\cref{ssec:baselines}). Our controller can handle complex (non-homogeneous) logical expressions.}
    \label{fig:detail}
\end{figure}

\section{Handling Complex Logic in Plans} 
\label{app:logic}
While the examples in \cref{fig:intro,fig:main} show homogeneous logic across sub-tasks in the plan, our controller can handle complex logical expressions including both ``\textsc{And}'' and ``\textsc{Or}'' operators. Specifically, we provide instructions to the planner to output this logical expressing at the end of the plan with a fixed prefix: \texttt{Execution Order}. We then build a deterministic parser that can parse complex logical expressions that the controller can process. We do so by splitting the logical expression into a series of homogeneous expression each passed to \method{}. Whenever the task given to \method{} comprises of multiple sub-tasks connected via (one) logical operator, we automatically decompose this task as per the logical expression. For example, in \cref{fig:detail}, a detailed plans used by the plan-and-execute baseline (discussed in \cref{ssec:baselines}) comprised of logical expressions using both \textsc{And}, and \textsc{Or} operators. Therefore, the parser will break automatically break this into multiple levels, i.e., Step 6 $=$ Step 1 \textsc{Or} Step 2 \textsc{Or} Step 3, followed by Step 6 \textsc{And} Step 4 \textsc{And} Step 5. While such complex logical expressions are mostly associated with the plan-and-execute baseline, they can be easily used within the \method{} framework. Furthermore, this allows the plan-and-execute baseline to simulate a multi-level planning structure via detailed plans without being adaptive to the executor.

\section{Task-specific Executors in ALFWorld}
\label{app:react}
In \cref{tab:alf}, we use a standardized executor with in-context demonstrations of atomic skills and two gold trajectories. While this allows for a common executor across different sub-tasks, task-specific executors yield higher performance on the specific sub-tasks. We now show \method{} can also be used on top of task-specific executors used by \citet{yao2023react}. The results are shown in \cref{tab:alt_react}. First, we observe that \method{} yields the overall success rate by up to $23.1\%$ points and also surpasses baselines on all but 1 task types. Interestingly, we find strong performance of the plan-and-execute baseline when using a stronger executor (as compared to \cref{tab:alf}) possibly as such an executor can handle complex sub-tasks better. Consistent with \cref{tab:alf}, \method{} outperforms Reflexion by $12.6\%$ points despite lack of dedicated memory and natural language feedback.

\begin{figure}[t]
    \centering
    \includegraphics[scale=0.5]{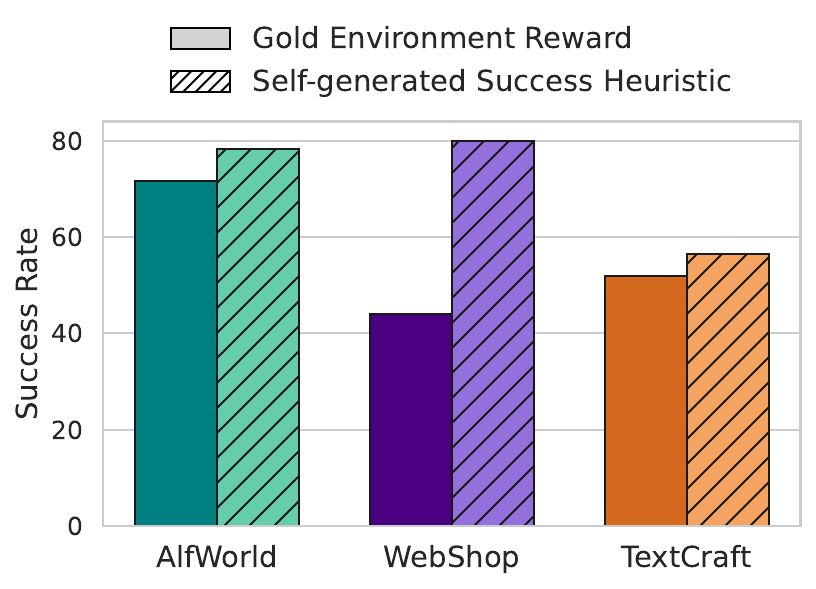}
    \caption{Comparison of LLM-generated success heuristic with gold environment rewards to compute success rates for all datasets.}
    \label{fig:eval}
\end{figure}
\section{Additional WebShop Experiments}
\label{app:web}

\paragraph{Evaluation Metrics.}
We focus on success rate and not the (soft) score as the primary metric for this task because it is possible to get a non-zero score by naively buying a product. To this effect, we construct a naive executor that inputs the user query in the search bar and buys the first available product. \cref{tab:web-score} shows that while this baseline yields the lowest success rate, it surprisingly yields a high success rate of 58.3. In contrast, our executors often do not buy products especially when the previous sub-goals fail which can adversely impact scores even though the success rate remains unaffected. Therefore, we argue for optimizing the success rate instead of the score as opposed to prior works~\cite{zhou2023language}.

\paragraph{\method{} accommodating task complexity.}

By default, \citet{yao2023react} use a search page with only the top-3 search results displayed. Intuitively, increasing the number of products on the search page requires the model to choose from a wider array of products and track all their information to determine the best fit to the user query, making the overall task harder. Therefore, we apply \method{} on Webshop in two settings with 3, and  10 products per search page.

\begin{table}[t]
\small
\centering
\setlength{\tabcolsep}{4pt}
\begin{tabular}{l c c }
    \toprule
      \bf Method  & \bf \#Products & \bf Success Rate \\
      \midrule
    ReAct & 3&  27.5\\
    \method{} ($d_{\mathrm{max}}=3$)  & 3 & \bf 47.5\\
    \midrule
    ReAct &10 &  20.0\\
    \method{} ($d_{\mathrm{max}}=3$) & 10 &  \bf{42.5}\\
    \bottomrule
    \end{tabular}
    \caption{\method{} improves WebShop (dev) performance irrespective of how many products (3 or 10) are chosen from the search page.}
    \label{tab:task}
\end{table}

\paragraph{Results.} From \cref{tab:task}, we observe that \method{} effectively improves success rate by $20.0\%$ and $22.5\%$ for 3 and 10 products respectively over the ReAct baseline. The difference in ReAct performance for both settings corroborates our hypothesis that increasing number of products on the search page increases task complexity, all else equal. Notably, we show that \method{} yields \emph{higher} improvement for \emph{more complex} task settings.
\section{TextCraft}
\label{app:textcraft}
\paragraph{TextCraft: Environment Details.} In TextCraft, the objective is to obtain target Minecraft items by crafting them from available items in the environment. We define an environment with three actions: \texttt{craft <item> using <ingredients>}, \texttt{get <item>}, and \texttt{inventory}. We utilize Minecraft's crafting recipes to specify craftable items and their ingredients, assuming that all other items are obtainable from the environment. Similar to AlfWorld, our agent can directly execute these operations in the embodied game. The game begins with a list of crafting commands provided to the agent that detail recipes that can be used to craft the final target, its ingredients along with some distractors (details in \cref{app:textcraft}). A reward of 1 is generated when the target item gets added to the agent's inventory. An illustrative gold trajectory from TextCraft is shown in \cref{fig:textcraft}.

We create the TextCraft environment using Minecraft v1.16.5 recipes. We only consider the recipes craftable using a crafting table. We consider both shapeless (only count matters) and shaped (position of ingredients matters) recipes and convert them into crafting commands (e.g. \texttt{craft 4 sticks using 2 planks}). Items that do not have any recipe are considering obtainable via the \texttt{get} command, e.g. \texttt{get 4 diamond}. 

Since the entire set of crafting commands would not fit in the context of modern LLMs, we create a set of relevant crafting commands for every task. Apart from the set of gold crafting commands (i.e, crafting commands for all the items in the recipe tree), we also add up to 10 distractor commands. To create this distractor set, we sub-sample up to 10 recipes for every ingredient in the recipes of our gold recipe tree. We finally sub-sample up to 10 distractors from this entire set to ensure a reasonable context size. Note that we do not provide the list of valid \texttt{get} commands as that can be inferred from the \texttt{craft} commands.

\section{Evaluation of  Success Heuristic}
\label{ssec:comm}
In \cref{ssec:exec}, we describe the executor module used in \method{}. For tasks assigned to the executor, we prompt the LLM to generate a binary success heuristic. We use this heuristic repeatedly to evaluate if the (sub-)task needs to be decomposed further. We now study the ability of LLMs to generate this success heuristic on all our datasets. To this end, we run \method{} and in the end compare the success rate when using the LLM's self-assessed task success with the gold reward from the environment in \cref{fig:eval}. On ALFWorld and TextCraft, we find the LLM slightly over-estimates its overall task success. This is to be expected as the underlying tasks involve minimal subjectivity (e.g., the agent either has an item on its inventory or not). However, on WebShop, where a product can match the user criteria to different degrees (partially or fully), we find that the LLM's assessment is significantly inflated compared to the environment reward ($>\!30$ points). This imperfect feedback affects downstream performance of \method{}, as the algorithm terminates even though further decomposition is needed. We leave it to future work to address the shortcomings of self-evaluation with LLMs~\cite{ huang2023large, stechly2023gpt4}.

\section{Prompts}
\label{app:prompts}
We provide all the prompts used in our planner and executor modules for ALFWorld, WebShop, and TextCraft datasets in the following pages.

\begin{figure*}
\centering
\begin{minipage}{0.95\textwidth}
\begin{lstlisting}[title=\texttt{ALFWorld Hybrid Executor Prompt},basicstyle=\ttfamily\scriptsize,backgroundcolor=\color{lavender}]
Here is a demo of actions you can perform.

You are in the middle of a room. Looking quickly around you, you see a desk 1, microwave 1, a cabinet 3, a cabinet 9, a drawer 2, a coffeemachine 1, a stoveburner 4, a drawer 5, a cabinet 11, a drawer 3, a stoveburner 1, a drawer 1, a toaster 1, a fridge 1, a stoveburner 2, a cabinet 6, a cabinet 10, a countertop 1, a cabinet 13, a cabinet 7, a garbagecan 1, a cabinet 2, a cabinet 8, a cabinet 12, a drawer 4, a cabinet 1, a sinkbasin 1, a cabinet 5, a stoveburner 3, and a cabinet 4.

Your task is to: Put a tomato in cabinet 1.
> think:  Input assumption: I am carrying a tomato. Now I need to verify this.
OK.
> inventory
You are carrying: a tomato 1.
> examine tomato 1
There's nothing special about tomato 1.
> think: Based on previous context assumption is true, now I can proceed.
> go to cabinet 1
The cabinet 1 is closed.
> open cabinet 1
You open the cabinet 1. The cabinet 1 is open. In it, you see a candle 1, and a spraybottle 2.
> put tomato 1 in/on cabinet 1
You put the tomato 1 in/on the cabinet 1.
>  think: End state: Now I put tomato (1) in cabinet (1). Task completed!

Your task is to: Take a mug from cabinet 5
> think: Input assumption: None
OK.
> go to cabinet 5
The cabinet 5 is closed.
> open cabinet 5
You open the cabinet 5. The cabinet 5 is open. In it, you see a mug 3.
> take mug 3 from cabinet 5
You pick up the mug 3 from the cabinet 5.
> think: End state: Now I take a mug (3) from cabinet (5). Task completed!

Your task is to: Heat an egg.
> think:  Input assumption: I am carrying an egg. Now I need to verify this.
OK.
> inventory
You are carrying: a egg 2.
> examine egg 2
There's nothing special about egg 2.
> think: Based on previous context assumption is true, now I can proceed.
OK.
> go to microwave 1
The microwave 1 is closed.
> heat egg 2 with microwave 1
You heat the egg 2 using the microwave 1.
> think: End state: Now I heat a egg (2). Task completed!

Your task is to: Cool a mug.
> think:  Input assumption: I am carrying a mug. Now I need to verify this.
OK.
> inventory
You are carrying: a mug 3.
> examine mug 3
There's nothing special about mug 3.
> think: Based on previous context assumption is true, now I can proceed.
OK.
> go to fridge 1
The fridge 1 is closed.
> cool mug 3 with fridge 1
You cool the mug 3 using the fridge 1.
> think: End state:  Now I cool a mug (3). Task completed!

Your task is to: Clean a lettuce.
> think:  Input assumption: I am carrying a lettuce. Now I need to verify this.
OK.
> inventory
You are carrying: a lettuce 1.
> examine lettuce 1
There's nothing special about lettuce 1.
> think: Based on previous context assumption is true, now I can proceed.
OK.
> go to sinkbasin 1
On the sinkbasin 1, you see a apple 2, a ladle 2, a spoon 1, and a tomato 3.
> clean lettuce 1 with sinkbasin 1
You clean the lettuce 1 using the sinkbasin 1.
> think: End state:  Now I clean a lettuce (1). Task completed!

\end{lstlisting}
\end{minipage}
\label{prmpt:alf_exec_1}
\end{figure*}
\begin{figure*}
\centering
\begin{minipage}{0.95\textwidth}
\begin{lstlisting}[title=\texttt{ALFWorld Hybrid Executor Prompt (cont.)},basicstyle=\ttfamily\scriptsize,backgroundcolor=\color{lavender}]
Your task is to: Use a desklamp to look at tomato 1.
> think: Input assumption: I am at the location that contains desklamp (1). Now I need to verify this.
OK.
> inventory
 You are carrying: a tomato 1.
> look
You are facing desk 1. Next to it, you see nothing.
> examine desk 1
On desk 1, you see a spoon 1, a desklamp 1, and a pencil 2.
> think: Based on previous context assumption is true, now I can proceed.
OK.
> use desklamp 1
You turn on the desklamp 1.
> think: Now I use desklamp (1). Task completed!

Here are two example tasks.

You are in the middle of a room. Looking quickly around you, you see a cabinet 13, a cabinet 12, a cabinet 11, a cabinet 10, a cabinet 9, a cabinet 8, a cabinet 7, a cabinet 6, a cabinet 5, a cabinet 4, a cabinet 3, a cabinet 2, a cabinet 1, a coffeemachine 1, a countertop 1, a diningtable 1, a drawer 1, a fridge 1, a garbagecan 1, a microwave 1, a shelf 3, a shelf 2, a shelf 1, a sinkbasin 1, a stoveburner 4, a stoveburner 3, a stoveburner 2, a stoveburner 1, and a toaster 1.
Your task is to: put a hot apple in fridge.
> think: To solve the task, I need to find and take an apple, then heat it with microwave, then put it in fridge.
OK.
> think: First I need to find an apple. An apple is more likely to appear in fridge (1), diningtable (1), coffeetable (1), drawer (1), cabinet (1-13), garbagecan (1). I can check one by one, starting with fridge 1.
OK.
> go to fridge 1
The fridge 1 is closed.
> open fridge 1
You open the fridge 1. The fridge 1 is open. In it, you see a cup 1, and a egg 1.
> go to diningtable 1
On the diningtable 1, you see a apple 1, a bread 1, a fork 3, a fork 2, a fork 1, a mug 2, a peppershaker 3, a plate 2, a pot 1, a soapbottle 3, a spatula 1, a spoon 3, a spoon 2, a spoon 1, a winebottle 3, and a winebottle 2.
> think: Now I find an apple (1). Next, I need to take it.
OK.
> take apple 1 from diningtable 1
You pick up the apple 1 from the diningtable 1.
> think: Now I take an apple (1). Next, I need to go to a microwave (1) and heat it.
OK.
> go to microwave 1
The microwave 1 is closed.
> heat apple 1 with microwave 1
You heat the apple 1 using the microwave 1.
> think: Now I heat an apple (1). Next, I need to put it in/on fridge 1.
OK.
> go to fridge 1
The fridge 1 is open. In it, you see a cup 1, and a egg 1.
> put apple 1 in/on fridge 1
You put the apple 1 in/on the fridge 1.

> think: Task completed!

You are in the middle of a room. Looking quickly around you, you see a bed 1, a desk 1, a drawer 3, a drawer 2, a drawer 1, a garbagecan 1, a safe 1, a shelf 5, a shelf 4, a shelf 3, a shelf 2, a shelf 1, a sidetable 2, and a sidetable 1.
Your task is to: look at bowl under the desklamp.
> think: To solve the task, I need to find and take a bowl, then find and use a desklamp. 
OK. 
> think: First I need to find a bowl. A bowl is more likely to appear in drawer  shelf (1-5), drawer (1-3). I can check one by one, starting with drawer 1.
OK.
> go to shelf 1
On the shelf 1, you see nothing.
> go to shelf 2
On the shelf 2, you see a bowl 1.
> think: Now I find a bowl (1). Next, I need to take it.
OK.
> take bowl 1 from shelf 2
You pick up the bowl 1 from the shelf 2.
> think: Now I take a bowl (1). Next, I need to find a desklamp. A desklamp is more likely to appear in desk (1), sidetable (1-2), and drawer (1-3). I can check one by one, starting with desk 1.
OK.
> go to desk 1
On the desk 1, you see a alarmclock 2, a book 1, a cellphone 1, a keychain 1, a laptop 2, a laptop 1, and a pen 3.
> go to sidetable 1
On the sidetable 1, you see a cd 1, a pen 1, and a pencil 1.
> go to sidetable 2
On the sidetable 2, you see a alarmclock 1, a desklamp 1, and a pen 2.
> think: Now I find a desklamp (1). Next, I need to use it.
OK.
> use desklamp 1
You turn on the desklamp 1.
> think: Task completed!
\end{lstlisting}
\end{minipage}
\label{prmpt:alf_exec_2}
\end{figure*}
\begin{figure*}
\centering
\begin{minipage}{0.95\textwidth}
\begin{lstlisting}[title=\texttt{ALFWorld Planner Prompt}, basicstyle=\ttfamily\scriptsize,backgroundcolor=\color{peach}]
Here are some examples.
You are in the middle of a room. Looking quickly around you, you see a desk 1, microwave 1, a cabinet 3, a cabinet 9, a drawer 2, a coffeemachine 1, a stoveburner 4, a drawer 5, a cabinet 11, a drawer 3, a stoveburner 1, a drawer 1, a toaster 1, a fridge 1, a stoveburner 2, a cabinet 6, a cabinet 10, a countertop 1, a cabinet 13, a cabinet 7, a garbagecan 1, a cabinet 2, a cabinet 8, a cabinet 12, a drawer 4, a cabinet 1, a sinkbasin 1, a cabinet 5, a stoveburner 3, and a cabinet 4.

Goal: Put a mug in/on desk.
Come up with an abstract plan to perform this task in a couple of steps.
# Think: To perform this task, I need to find and take mug and then put it on desk. First, I will focus on finding mug.
Step 1: Find and take mug
# Think: Now that I am carrying mug, I will focus on putting it in/on desk.
Step 2: Put mug in/on desk
Execution Order: (Step 1 AND Step 2)

Goal: Clean mug and put it in/on desk.
Come up with an abstract plan to perform this task in a couple of steps.
# Think: To perform this task, I need to find and take mug, clean it, and then put it on desk. First, I will focus on finding mug.
Step 1: Find and take mug
# Think: Now that I am carrying mug, I will focus on cleaning it.
Step 2: Clean mug with sinkbasin
# Think: Now that I have cleaned mug, I will focus on putting it in/on desk.
Step 3: Put cleaned mug in/on desk
Execution Order: (Step 1 AND Step 2 AND Step 3)

Goal: Cool mug and put it in/on desk.
Come up with an abstract plan to perform this task in a couple of steps.
# Think: To perform this task, I need to find and take mug, cool it, and then put it on desk. First, I will focus on finding mug.
Step 1: Find and take mug
# Think: Now that I am carrying mug, I will focus on cooling it.
Step 2: Cool mug with fridge
# Think: Now that I have cooled mug, I will focus on putting it in/on desk.
Step 3: Put cooled mug in/on desk
Execution Order: (Step 1 AND Step 2 AND Step 3)

Goal: Heat mug and put it in/on desk.
Come up with an abstract plan to perform this task in a couple of steps.
# Think: To perform this task, I need to find and take mug, heat it, and then put it on desk. First, I will focus on finding mug.
Step 1: Find and take mug
# Think: Now that I am carrying mug, I will focus on heating it.
Step 2: Heat mug with microwave
# Think: Now that I have heated mug, I will focus on putting it in/on desk.
Step 3: Put heated mug in/on desk
Execution Order: (Step 1 AND Step 2 AND Step 3)

Goal: Look at mug under desklamp.
Come up with an abstract plan to perform this task in a couple of steps.
# Think: To perform this task, I need to find and take mug, and then go to the desklamp and use it. First, I will focus on finding mug.
Step 1: Find and take mug
# Think: Now that I have found and taken mug, I will focus on using the desklamp.
Step 2: Use the desklamp
Execution Order: (Step 1 AND Step 2)

Goal: Find and take mug
Come up with an abstract plan to perform this task in a couple of steps.
# Think: To perform this task I need to find mug in the room. mug is likely to be in desk, cabinet, countertop, or drawer. Now I will focus on finding mug in each of these locations one by one.
Step 1: Find and take mug from desk
# Think: If mug not found so far, I will next look in the cabinet.
Step 2: Find and take mug from cabinet
# Think: If mug not found so far, I will next look in the countertop.
Step 3: Find and take mug from countertop
# Think: If mug not found so far, I will next look in the drawer.
Step 4: Find and take mug from drawer
Execution Order: (Step 1 OR Step 2 OR Step 3 OR Step 4)

Here is the goal.
<room>
Goal: <task>.
Come up with an abstract plan to perform this task in a couple of steps. Constraints: The robot can hold/take/put only one object at a time to a location.
Ensure each step can be understood independently and mentions the name of object.
When stating the execution order, ensure that 'AND'/'OR' statements are properly nested using brackets '()'.
\end{lstlisting}
\end{minipage}
\label{prmpt:alf_plan}
\end{figure*}
\begin{figure*}
\centering
\begin{minipage}{0.95\textwidth}
\begin{lstlisting}[title=\texttt{WebShop Executor Prompt: Buy}, basicstyle=\ttfamily\scriptsize,backgroundcolor=\color{lavender}]
Instruction: Buy product [B078GWRC1J] that matches criteria: 3 ounce bottle of bright citrus deodorant for sensitive skin, and price lower than 50.00 dollars 

Action: think['I should be on the search page where I can see a list of products, I should verify that.']
Observation: OK.

Action: load
Observation:
[Back to Search] 
Page 1 (Total results: 50) 
[Next >] 
[B078GWRC1J] 
Bright Citrus Deodorant by Earth Mama | Safe for Sensitive Skin, Pregnancy and Breastfeeding, Contains Organic Calendula 3-Ounce 
$10.99 
[B08KBVJ4XN] 
Barrel and Oak - Aluminum-Free Deodorant, Deodorant for Men, Essential Oil-Based Scent, 24-Hour Odor Protection, Cedar & Patchouli Blend, Gentle on Sensitive Skin (Mountain Sage, 2.7 oz, 2-Pack) 
$35.95 
[B078GTKVXY] 
Ginger Fresh Deodorant by Earth Mama | Natural and Safe for Sensitive Skin, Pregnancy and Breastfeeding, Contains Organic Calendula 3-Ounce 
$10.99 

Action: think['I can spot product [B078GWRC1J] and it is a clickable link since its enclosed in square brackets. I should click on it.]
Observation: OK.

Action: click[B078GWRC1J]
Observation:
[Back to Search] 
[< Prev] 
scent [assorted scents][bright citrus][calming lavender][ginger fresh][simply non-scents]
size [travel set (4-pack)][3 ounce (pack of 1)][3-ounce (2-pack)]
Bright Citrus Deodorant by Earth Mama | Natural and Safe for Sensitive Skin, Pregnancy and Breastfeeding, Contains Organic Calendula 3-Ounce 
Price: $10.99 
Rating: N.A. 
[Description] 
[Features] 
[Reviews] 
[Buy Now]

Action: think[I'm already on the product page.  I see multiple possible options enclosed in '[]' for scent and size for this product. To match the input criteria, I should click on options '[bright citrus]' for scent and '[3 ounce (pack of 1)]' for size one by one and then buy in the end.]
Observation: OK.

Action: click[bright citrus]
Observation: You have clicked bright citrus. 

Action: click[3 ounce (pack of 1)]
Observation: You have clicked 3 ounce (pack of 1). 

Action: think[My task is to buy the product, for it should to click 'buy now']
Observation: OK.

Action: click[Buy Now]
Observation: You have clicked buy now.

Action: think[I finished buying the product. Task completed!]


Here is another task in which you need to buy a product. When you finish buying the product with the most relevant choices, use 'think[Task completed']. If you cannot find the matching options or proceed, think['Task failed']. Note that you can only click on text enclosed in '[]' on the webpage. Everything else is only a description, not valid with the "click" action.

Instruction: Buy product [{}] that matches the criteria: {}



\end{lstlisting}
\end{minipage}
\label{prmpt:web_exec_1}
\end{figure*}
\begin{figure*}
\centering
\begin{minipage}{0.95\textwidth}
\begin{lstlisting}[title=\texttt{WebShop Executor Prompt: Match (cont.)}, basicstyle=\ttfamily\scriptsize,backgroundcolor=\color{lavender}]
You are given a webpage of an item on an online shopping website and a criteria. Your task is to answer if the product on the page exactly matches the criteria. Not the criteria could have multiple requirements that should be checked one by one and all must satisfy for an exact match.

Here are a few examples:

Criteria: 3 ounce bottle of citrus deodorant for sensitive skin that is priced lower than $30 and natural.
Item Page:
[Back to Search] 
[< Prev] 
scent [assorted scents][bright citrus][calming lavender][ginger fresh][simply non-scents]
size [travel set (4-pack)][3 ounce (pack of 1)][3-ounce (2-pack)]
Bright Citrus Deodorant by Earth Mama | Natural and Safe for Sensitive Skin, Pregnancy and Breastfeeding, Contains Organic Calendula 3-Ounce 
Price: $10.99 
Rating: N.A. 
[Description] 
Features:
 NEW from Earth Mama (formerly Earth Mama Angel Baby), formulated especially for pregnancy, breastfeeding and sensitive skin  
 Contains organic grapefruit, tangerine and calendula  
 NO propylene glycol, artificial fragrance, parabens or aluminum  
 Dermatologist tested and clinically tested for irritation  
 Better than natural organic! NSF/ANSI 305 Certified by Oregon Tilth   
[Reviews] 
[Attributes] 
[Buy Now] 

Answer: The product is available in 3 ounce size, is citrus and suitable for sensitive skin. It is also organic or natural. Its price is $10.99 which is less than $30.
Thus, the answer is True (exact match).

Criteria: 3 ounce bottle of citrus deodorant for sensitive skin that is priced lower than $30 and natural.
Item Page:
[Back to Search] 
[< Prev] 
size [3 ounce][3 ounce (pack of 1)]
unit count [2.0][3.0]
Barrel and Oak - Aluminum-Free Deodorant, Deodorant for Men, Essential Oil-Based Scent, 24-Hour Odor Protection, Cedar & Patchouli Blend, Gentle on Sensitive Skin (Mountain Sage, 2.7 oz, 2-Pack) 
Price: $15.95 
Rating: N.A. 
[Description] 
Features: 
About this item WHY ALUMINUM-FREE DEODORANT? Aluminum-free deodorants use more natural ingredients unlike antiperspirants, which use chemicals to block sweat. Safely fight odor for 24 hours with Barrel & Oak's deodorantsour gentle formula is easy on sensitive skin. START SMELLING LIKE THE MAN YOU WANT TO BE: Our mountain sage aluminum-free men's deodorant is naturally fragranced with an outdoorsy scent of crisp conifer, sage, & citrus. Think sweet notes of citrus with earthy tones of cedar & patchouli. PREMIUM INGREDIENTS FOR NATURAL FRAGRANCES: Our deodorants for men are composed of natural, essential oil-based scents. These natural fragrance deodorants are more subtle than their synthetic counterparts, but they're better for you & the planet. DESIGNED FOR THE MODERN MAN: Barrel & Oak has a full spectrum of grooming & body care products that are designed with function, fragrance, & effective ingredients for the health-conscious & practical modern man. Give your body what it deserves. EARTH-FRIENDLY, YOU-FRIENDLY, WALLET-FRIENDLY: Our premium products for men are scented with natural fragrances & essential oils, free of parabens, phthalates, & SLS, packaged in recyclable materials, cruelty-free, & vegan or vegetarian.
[Reviews] 
[Attributes] 
[Buy Now] 

Answer: The product is not citrus in nature. It does not match the criteria. It's price is $15.95 which is less than $30.
Thus, the answer is False (not an exact match).

Now here is the criteria and item page for the another task. Try you best to determine exact match, otherwise, respond with "False", i.e., no exact match. Generate an explanation before the answer to justify your decision.

Criteria: {}
Item Page:
{}
Answer: 
\end{lstlisting}
\end{minipage}
\label{prmpt:web_exec_2}
\end{figure*}
\begin{figure*}
\centering
\begin{minipage}{0.95\textwidth}
\begin{lstlisting}[title=\texttt{WebShop Executor Prompt: Shortlist (cont.)}, basicstyle=\ttfamily\scriptsize,,backgroundcolor=\color{lavender}]
You are given a search page on an online shopping site with a list of products along with name and price. Based on this information, your task is return a list of product IDs (enclosed in []) of all products that exactly match all requirements in the criteria. If the information provided is not enough to make a determination, return an empty list. 
Here are a few examples.

Search Page: 
[Back to Search] 
Page 1 (Total results: 50) 
[Next >] 
[B078GWRC1J] 
Bright Citrus Deodorant by Earth Mama | Safe for Sensitive Skin, Pregnancy and Breastfeeding, Contains Organic Calendula 3-Ounce 
$10.99 
[B08KBVJ4XN] 
Barrel and Oak - Aluminum-Free Deodorant, Deodorant for Men, Essential Oil-Based Scent, 24-Hour Odor Protection, Cedar & Patchouli Blend, Gentle on Sensitive Skin (Mountain Sage, 2.7 oz, 2-Pack) 
$35.95 
[B078GTKVXY] 
Ginger Fresh Deodorant by Earth Mama | Natural and Safe for Sensitive Skin, Pregnancy and Breastfeeding, Contains Organic Calendula 3-Ounce 
$10.99 
[B08SMG4WB9] 
Each & Every 2-Pack Natural Aluminum-Free Deodorant for Sensitive Skin with Essential Oils, Plant-Based Packaging (Citrus & Vetiver, 2.5 Ounce (Pack of 2)) 
$25.0 
[B08KVCCSD6] 
Each & Every 3-Pack, Natural Aluminum-Free Deodorant for Sensitive Skin Made with Essential Oils, 2.5 Oz. (Lavender & Lemon, Citrus & Vetiver, and Coconut & Lime) 
$35.0 

Criteria: less than 5 ounce citrus deodorant sensitive skin, price less than $30.
Answer: My requirements are 5 ounce, citrus deodrant, suitable for sensitive skin, and price less than $30. Looks like this information is available on the search page, so I can proceed.
Products B078GWRC1J, B08SMG4WB9 look suitable as they are less than 5 ounce, citrus and have price 10.99 and $25 less than $30. Thus, shortlisted IDs are shortlisted=['B078GWRC1J', 'B08SMG4WB9']

Criteria: less than 5 ounce citrus deodorant sensitive skin, cruelty free.
Answer: My requirements are 5 ounce, citrus deodrant, suitable for sensitive skin, and cruelty-free. Since there is no information about cruelty free on the search page, I cannot proceed. Task failed!

Here is another task with a different search page and criteria. List all the product ids (enclosed in []) from the search page that match ALL the requirements in the criteria. Name this list shortlisted. If you cannot make the determination about even 1 sub-criteria, do not make a guess, output "task failed!". Generate an explanation before the answer to justify your decision.

Search Page:
{}

Criteria: {}
Answer: 
\end{lstlisting}
\end{minipage}
\label{prmpt:web_exec_3}
\end{figure*}
\begin{figure*}
\centering
\begin{minipage}{0.95\textwidth}
\begin{lstlisting}[title=\texttt{WebShop Planner Prompt}, basicstyle=\ttfamily\scriptsize,backgroundcolor=\color{peach}]
Write an abstract plan to successfully complete the goal. In each step of the plan mention which module (including arguments) that need to be called. Learn from and incorporate information from previous runs, e.g. do not repeat previously successful or unsuccesful commands. Here are some examples:Information from previous run: -
Goal: Buy 3 ounce bottle of citrus deodorant for sensitive skin, that is natural and priced less than 50.00 dollars.
# Think: Based on the criteria and the search bar, I should query 3 ounce citrus deodorant sensitive skin. I have the following constraints: natural and price lower than $30 which I can use to narrow down search results.
Step 1: Search[3 ounce citrus deodorant sensitive skin]
# Think: Now I will need to narrow down the search results for price lower than $30 and natural
Step 2: SimpleMatch[3 ounce citrus deodorant sensitive skin with price lower than $50 and natural]
# Think: Since it returns a list of up to 3 products, I will pick the first suitable product. For now, Ill denote its id as prod_id for placeholder.
Step 3: Buy[prod_id, "3 ounce bottle of citrus deodorant for sensitive skin, that is natural and priced less than 30.00 dollars"]
#Think: My plan requrires all these steps to succeed sequentially, so I will use the "AND" operator.
Execution Order: (Step 1 AND Step 2 AND Step 3)


Information from previous run: 
- Unable to get matching product using: SimpleMatch[3 ounce citrus deodorant sensitive skin with price lower than $30 and natural]
- Search results page:
[Back to Search] 
Page 1 (Total results: 50) 
[Next >] 
[B078GWRC1J] 
Bright Citrus Deodorant by Earth Mama | Safe for Sensitive Skin, Pregnancy and Breastfeeding, Contains Organic Calendula 3-Ounce 
$10.99 
[B08KBVJ4XN] 
Barrel and Oak - Aluminum-Free Deodorant, Deodorant for Men, Essential Oil-Based Scent, 24-Hour Odor Protection, Cedar & Patchouli Blend, Gentle on Sensitive Skin (Mountain Sage, 2.7 oz, 2-Pack) 
$35.95 
[B078GTKVXY] 
Ginger Fresh Deodorant by Earth Mama | Natural and Safe for Sensitive Skin, Pregnancy and Breastfeeding, Contains Organic Calendula 3-Ounce 
$10.99 
[B08SMG4WB9] 
Each & Every 2-Pack Natural Aluminum-Free Deodorant for Sensitive Skin with Essential Oils, Plant-Based Packaging (Citrus & Vetiver, 2.5 Ounce (Pack of 2)) 
$25.0 
[B08KVCCSD6] 
Each & Every 3-Pack, Natural Aluminum-Free Deodorant for Sensitive Skin Made with Essential Oils, 2.5 Oz. (Lavender & Lemon, Citrus & Vetiver, and Coconut & Lime) 
$35.0 
[B087WKSR2G] 

Goal: Narrow down search results for 3 ounce bottle of citrus deodorant for sensitive skin that is priced lower than $30 and natural. You cannot search again.
#Think: Based on the search results and previous information, SimpleMatch failed because my criteria was too complex. Price constraint is easy to verify, I will narrow down based on that first then examine in detail for natural constraint
#Think: Based on price, I narrow down my search to B078GWRC1J, B08SMG4WB9 as they look suitable. These are on my shortlist to examine the natural constraint  in detail one by one.
Step 1: DetailMatch[B078GWRC1J, 3 ounce bottle of  for sensitive skin, that is natural and priced less than 30.00 dollars]
Step 2: DetailMatch[B08SMG4WB9, 3 ounce bottle of citrus deodorantcitrus deodorant for sensitive skin, that is natural and priced less than 30.00 dollars]
#Think: If none of the products exactly match my criteria, I will search again with a new query that includes the natural criteria too. This ensures my plan is compelete.
Step 3: Search[3 ounce citrus deodrant natural and sensitive skin]
#Think: Since these steps are linked by an if condition, I only need one of them to succeed. I will connect them using the "OR" operator.
Execution Order: (Step 1 OR Step 2 OR Step 3)


Here is a new goal. Write an abstract plan to successfully complete the goal. In each step of the plan mention which module (including arguments) that need to be called. Learn from and incorporate information from previous runs, e.g. do not repeat previously successful or unsuccesful commands. In the end, output the intended execution order.

Information from previous run: {}
Goal: {}

\end{lstlisting}
\end{minipage}
\label{prmpt:web_plan}
\end{figure*}

\begin{figure*}
\centering
\begin{lstlisting}[title=\texttt{TextCraft Executor Prompt}, basicstyle=\ttfamily\scriptsize,backgroundcolor=\color{lavender}]
You are given few useful crafting recipes to craft items in Minecraft. Crafting commands are of the format "craft [target object] using [input ingredients]". You can either "fetch" an object (ingredients) from the inventory or the environment or "craft" (target) using any of the crafting commands. You can use ONLY these crafting commands provided, do not use your own crafting commands. However, if the crafting command uses a generic ingredient like "planks", you can use special types of the same ingredient e.g. "dark oak planks" in the command instead. For any other natural language or thoughts, use prefix 'think: '.

Here is a demo of how to fetch and craft objects.
Crafting commands:
craft 3 dark oak sign using 6 dark oak planks, 1 stick
craft 4 dark oak planks using 1 dark oak log
craft 1 stick using 1 planks
craft 4 stick using 2 bamboo
craft 4 oak planks using 1 oak log
craft 1 dark oak fence using 2 stick, 4 dark oak planks
craft 1 warped stairs using 6 warped planks
craft 3 oak sign using 6 oak planks, 1 stick
Goal: craft dark oak sign
> think: I should check if I can fetch dark oak sign directly from the environment or the inventory.
OK.
> inventory: 
Inventory: [stick] (1) [dark oak planks] (8)
> get dark oak sign
Could not find dark oak sign
> think: I cannot get dark oak sign directly, I need to craft it. From the crafting commands, I can use: craft dark oak sign using 6 dark oak planks, 1 stick. Ingredients needed: 6 dark oak planks, 1 stick. Input assumption: I have all the neccessary ingredients in my inventory. Let me verify this first.
OK.
> inventory
Inventory: [stick] (1) [dark oak planks] (8)
> think: I found my ingredients: 6 dark oak planks, 1 stick in my inventory. My assumption is true, I can proceed. I will use the crafting command: craft dark oak sign using 6 dark oak planks
OK.
> craft 1 dark oak sign using 6 dark oak planks, 1 stick
Crafted 1 minecraft:dark_oak_sign
> inventory 
Inventory: [dark oak sign] (1)
> think: I now have dark oak sign in my inventory. Task Completed!
OK.

Goal: fetch 2 dark oak logs.
> think: I should check my inventory first, to see if I already have dark oak sign. Otherwise, I will directly try to get it from the environment.
OK.
> inventory
Inventory: [stick] (1)
> get 2 dark oak logs.
Got 2 dark oak logs
> inventory
Inventory: [dark oak log] (2) [stick] (1)
> think: I have 2 dark oak logs in my inventory. Task Completed!
OK.

\end{lstlisting}
\end{figure*}

\begin{figure*}
\centering
\begin{lstlisting}[title=\texttt{TextCraft Executor Prompt (cont.)}, basicstyle=\ttfamily\scriptsize,backgroundcolor=\color{lavender}]
Here is an example of a complex goal.
Goal: craft 2 oak planks
> think: I should check if I can fetch dark oak sign directly from the environment or the inventory.
OK.
> inventory: 
Inventory: [stick] (1) 
> get 2 oak planks
Could not find 2 oak planks
> think: I cannot get oak planks directly, I need to craft it. From the crafting commands, I can use: craft 4 oak planks using 1 oak log. Ingredients needed: 1 dark oak log. Input assumption: I have the ingredients needed in my inventory. Let me verify this first.
OK.
> inventory
Inventory: [stick] (1) 
> think: My assumption is false. I need to get the ingredients: 1 oak log first.
OK.
> get 1 oak log
Got 1 oak log
> craft 4 oak planks using 1 oak log
Crafted 4 minecraft:oak_planks
> inventory
Inventory: [stick] (1) [oak_planks] (4)
> think: I now have 2 oak planks in my inventory. Task Completed!
OK.
Now here is a different goal. You can use these crafting commands to accomplish the goal. When you the desired item in your inventory, think: Task Completed! If you have tried your best but cannot proceed, think: task failed!
\end{lstlisting}
\end{figure*}

\begin{figure*}
\centering
\begin{minipage}{0.95\textwidth}
\begin{lstlisting}[title=\texttt{TextCraft Planner Prompt}, basicstyle=\ttfamily\scriptsize,backgroundcolor=\color{peach}]
Your task is to come up with a short plan to help me accomplish my goal in a couple of steps using at most ONE of the provided crafting commands. You can take the help of crafting commands below to create new objects. 
Craft command can be understood as follows: craft [target] using [ingredients], where target is item/object generated by the craft command as output and ingredient are the inputs. You are given an agent that can "craft" or "fetch" objects.

Here is are some examples.

Crafting commands:
craft 3 dark oak sign using 6 dark oak planks, 1 stick
craft 4 dark oak planks using 1 dark oak log
craft 1 stick using 1 planks
craft 4 stick using 2 bamboo
craft 4 oak planks using 1 oak log
craft 1 dark oak fence using 2 stick, 4 dark oak planks
craft 1 warped stairs using 6 warped planks
craft 3 oak sign using 6 oak planks, 1 stick

Goal: craft dark oak sign.

# Think: My target is a dark oak sign. From the list of crafting commands, only 1 command generates my target: craft 3 dark oak sign using 6 oak planks, 1 stick. I will use this command to devise a plan. My ingredients are: 6 dark oak planks, 1 stick. I should first get all the ingredients and then use the crafting command.
Step 1: fetch 6 dark oak planks
Step 2: fetch 1 stick
# Think: Now that I have collected the input ingredients, I can craft the dark oak sign using given command.
Step 3: craft dark oak sign using 6 dark oak planks, 1 stick
# Think: To succeed, I need to perform all these steps, one after the other. So I need to use the "AND" operator.
Execution Order: (Step 1 AND Step 2 AND Step 3)

Goal: fetch 6 dark oak planks.

# Think: My target is 6 dark oak planks. From the list of crafting commands, only 1 command generates my target: craft 4 dark oak planks using 1 dark oak log. My ingredients are: 1 dark oak log. To successfully accomplish the goal, I should first get all the ingredients and then use the crafting command.
Step 1: fetch 1 dark oak log
# Think: Now that I have collected the input ingredients, I can craft dark oak planks using given command. I know that I cannot use a partial recipe.
Step 2: craft 4 dark oak planks using 1 dark oak log
# Think: This gives me 4 dark oak planks which is less than my desired 6 dark oak planks. I know that I cannot use a partial recipe. So my goal is not satisfied, I need to craft more dark oak planks by repeating Step 2 one more time.
Step 3: craft 4 dark oak planks using 1 dark oak log
# Think: To succeed, I need to perform all these steps, one after the other. So I need to use the "AND" operator.
Execution Order: (Step 1 AND Step 2 AND Step 3)

Here is a different goal with different craft commands. Your task is to come up with a short plan to help me accomplish my goal in a couple of steps using at most ONE of the provided crafting commands. You can take the help of crafting commands below to create new objects. Keep in mind that:
- It is okay to generate more target objects than your goal.
- Be very careful with the count of objects, SAME object counts mentioned in the input crafting command. 
- You cannot use a partial crafting command recipe, i.e. if the recipe generates 2 objects you CANNOT alter it to produce just 1. 
- Also, you can use ONLY 1 crafting command in your plan.

\end{lstlisting}
\end{minipage}
\label{prmpt:text_plan}
\end{figure*}

\end{document}